\documentclass[conference,a4paper,10pt]{IEEEtran}
\usepackage{balance}
\usepackage{subcaption}
\usepackage{color}
\usepackage{graphics,graphicx}
\graphicspath{{./fig/}{./images/}}

\ifCLASSINFOpdf
\else
\fi
\usepackage{amsmath,amssymb}
\usepackage{algorithm}
\usepackage{algorithmic}
\usepackage{url}
\begin{document}
%\title{Bare Demo of IEEEtran.cls for Conferences}

\title{An Occlusion Reasoning Scheme for Monocular Pedestrian
Tracking in Dynamic Scenes}

\author{\IEEEauthorblockN{Sourav Garg and Swagat Kumar}
\IEEEauthorblockA{
Innovation Lab\\
Tata Consultancy Services \\
New Delhi, India 201301\\
Email: \{swagat.kumar,garg.sourav\}@tcs.com}
\and
\IEEEauthorblockN{Rajesh Ratnakaram and Prithwijit Guha}
\IEEEauthorblockA{Department of Electronics and Electrical Engineering \\
  Indian Institute of Technology Guwahati\\
Guwahati, Assam, India 781039\\
Email: \{pguha,ratnakaram\}@iitg.ernet.in}}


%Telephone: (800) 555--1212, Fax: (888) 555--1212}


\maketitle

\begin{abstract}

  This paper looks into the problem of pedestrian tracking  using a
  monocular, potentially moving, uncalibrated camera. The pedestrians are
  located in each frame using a standard human detector, which are then
  tracked in subsequent frames. This is a challenging problem as one
  has to deal with complex situations like changing background,
  partial or full occlusion and camera motion. In order to carry out
  successful tracking, it is necessary to resolve associations between
  the detected windows in the current frame with those obtained from
  the previous frame. Compared to methods that use temporal windows
  incorporating past as well as future information, we attempt to make
  decision on a frame-by-frame basis. An occlusion reasoning scheme is
  proposed to resolve the association problem between a pair of
  consecutive frames by using an affinity matrix that defines the
  closeness between a pair of windows and then, uses a binary integer
  programming to obtain unique association between them. A second
  stage of verification based on SURF matching is used to deal with
  those cases where the above optimization scheme might yield wrong
  associations. The efficacy of the approach is demonstrated through
  experiments on several standard pedestrian datasets.
\end{abstract}


% \fi
\IEEEpeerreviewmaketitle

%\IEEEkeywords{Pedestrian Detection and Tacking, Dynamic Scenes, Histogram
%of Gradients, Extended HoG, Occlusion resolution, Kalman Filter
%Prediction}

\section{Introduction} \label{sec:intro}

In this paper, we look into the problem of tracking multiple targets
using a monocular, possibly moving, uncalibrated camera. It has
several applications in areas like smart vehicles, robotics and video
surveillance. It can be used for extracting higher level of
information from a video, such as, event detection, crowd analysis
etc. The task involves locating concerned targets, assigning unique
IDs to each one of them and generating trajectories for them. 
The problem is challenging as one has to deal with several complex
situations like changing background, camera motion, wide variation in
appearance and illumination and, partial or full occlusion. 

One of the popular approach is to use 
tracking-by-detection framework which has become one of the popular approach
to solve this problem. In this framework, a detector is used to locate
targets in each frame and then associate these detections across
frames. This approach however suffers from the limitations of the
object detector which may yield false positives and missing
detections. On the other hand, resolving associations between detected
targets across frames may become challenging under conditions of group
formation and occlusions for long duration. 

In most of the methods, the data association problem is solved by
optimizing the detection assignments over a temporal window
\cite{perera2006multi} \cite{wu2007detection}
\cite{andriluka2008people}. Nevatia's group, particularly, focused on
hierarchical association at multiple levels \cite{wu2007detection}
\cite{huang2008robust} \cite{li2009learning} where the tracklets are
associated to form longer trajectories. The association is formulated
as a MAP problem which is solved using the Hungarian algorithm. These were
mostly off-line approaches where the frames were revisited over
multiple iterations.  In their latest work \cite{yang2014multi},
authors learn a Conditional Random Field (CRF) model to learn
appearance and motion model that also takes into account relative
positions between the targets. There are other approaches that make
use of particle filter to solve the tracking problem as in
\cite{okuma2004boosted} \cite{breitenstein2009robust}.

\begin{figure}[!t]
  \begin{center}
    \begin{tabular}{cccc}
      \rotatebox{90}{Without ORS} & 
      \includegraphics[width=2cm,height=3cm]{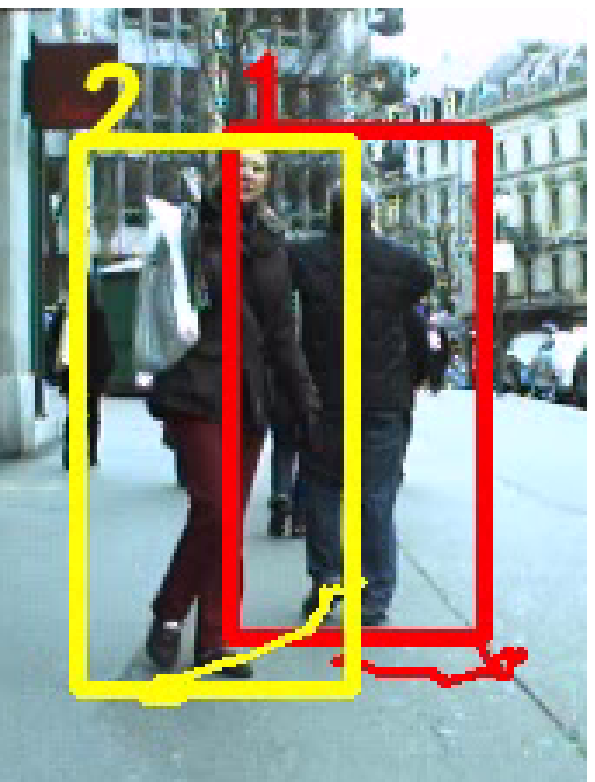} & 
      \includegraphics[width=2cm,height=3cm]{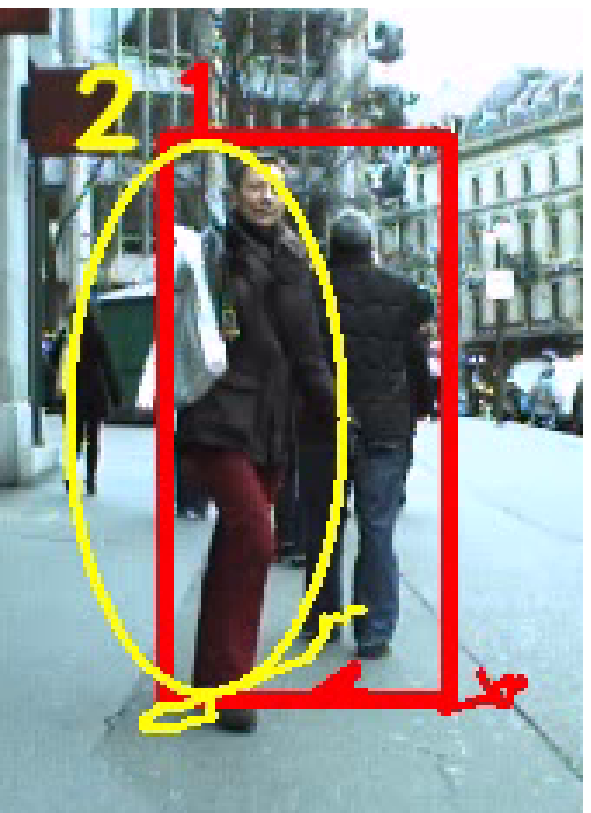} &  
      \includegraphics[width=2cm,height=3cm]{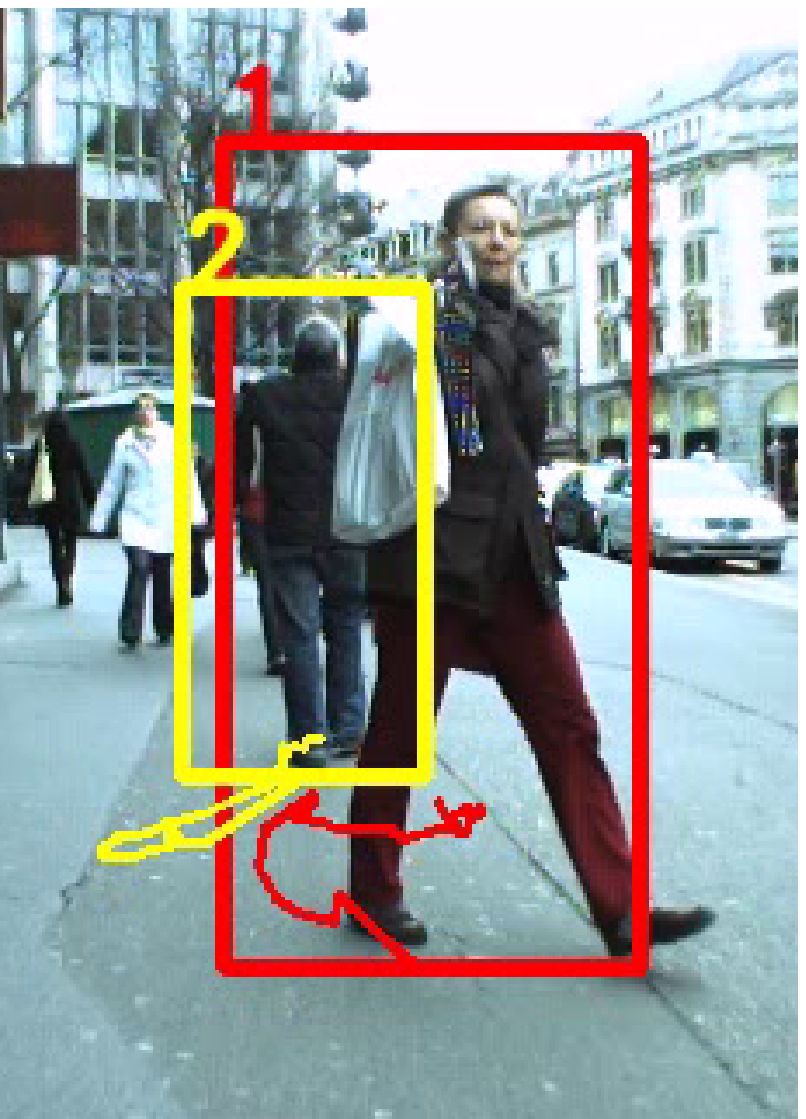} \\
      \rotatebox{90}{With ORS} & 
      \includegraphics[width=2cm,height=3cm]{before_occ} & 
      \includegraphics[width=2cm,height=3cm]{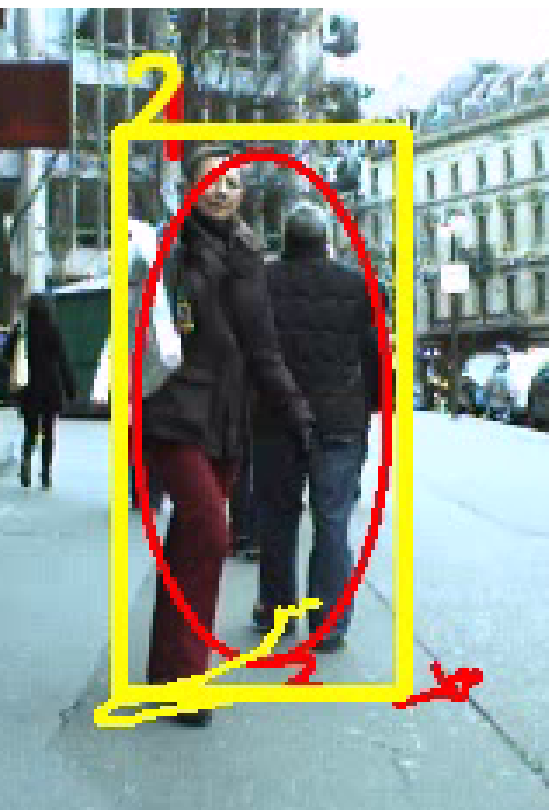} & 
      \includegraphics[width=2cm,height=3cm]{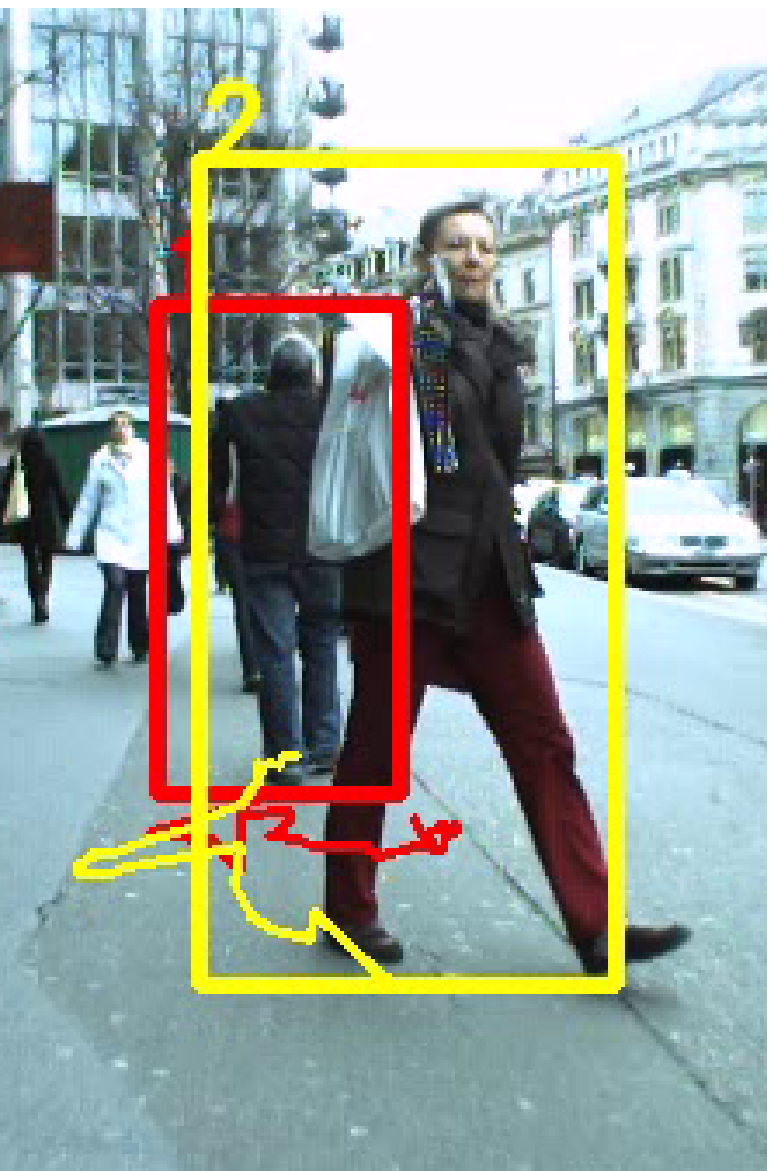} \\
      & Before Occlusion & Occlusion & After Occlusion
    \end{tabular}
  \end{center}
  \caption{Need for Occlusion Reasoning Scheme (ORS) in multi-agent
  tracking. Without ORS, agent IDs are interchanged between the
overlapping agents 1 and 2. This ID switch is prevented using ORS. }
  \label{fig:ors_need}
\end{figure}

In this paper, we re-look at the multi-target tracking problem with a
focus on simplifying the entire approach. We primarily focus on using
monocular camera images in contrast to other methods that use
stereo-vision system \cite{ess2008mobile} \cite{keller2011benefits},
laser scanner \cite{fuerstenberg2002pedestrian}, night vision
\cite{xu2005pedestrian} or LIDAR \cite{szarvas2006real}
\cite{premebida2007lidar}, sometimes in addition to vision.   
We aim to make online decisions on a frame-by-frame basis
unlike other approaches where a temporal window is used for
incorporating future information for resolving association in the
current frame \cite{yang2014multi}.  Such methods are prone to
frequent ID switches and trajectory fragmentation due to noisy and
ambiguous observation \cite{li2009learning}. We attempt to overcome
these limitations of a frame-by-frame approach in this paper. We use
the standard JRoG detector as used by the authors in
\cite{yang2014multi} as we are primarily making comparison with their
results. However, any other object detector could be used locating
pedestrians in the video. Readers may refer to
\cite{enzweiler2009monocular} \cite{geronimo2010survey}
\cite{dollar2012pedestrian} for a survey on the state of the art
methods in pedestrian detection. Once a new person is detected in a
frame, a colour-based mean-shift tracker is initialized. This
mean-shift (MS) tracker \cite{Comaniciu:2003} combined with a Kalman
Filter (KF) \cite{Haykin:2009} based motion predictor is used to
localize this target in the new frame. In order to carry out
successful tracking, it is necessary to resolve association between
the currently detected target windows with those estimated from the
previous frame using KF and MS tracker. This is challenging as the
windows may overlap with each other resulting in many-to-one or
one-to-many associations. The need for an occlusion reasoning scheme
in a multi-agent tracking problem is illustrated in Figure
\ref{fig:ors_need}. It is shown that the agent IDs get interchanged
during occlusion when the associations are not  properly and hence,
there is a need for having an effective occlusion reasoning
scheme.                         

Our main contribution lies in proposing an occlusion reasoning scheme
(ORS) that uses an affinity matrix and binary integer programming to
resolve the data association problem between a pair of frames. The
affinity matrix represents the `closeness' between a pair of tracking
windows. The binary integer programming (BIP) module returns unique
associations between the agents of this pair. Since the resolution is
based on some feature based scalar value function in the affinity
matrix, the resulting association might still be incorrect in some
extreme cases. We use agent pairing information from the last frame
and SURF matching to provide a second stage of verification over the
decision obtained from the BIP module. The resulting algorithm is
tested on several datasets and the performance is compared with the
state of the art.  

It is to be noted that one can use Hungarian algorithm
\cite{perera2006multi} \cite{yang2014multi} in place of BIP for
resolving associations. However, we envisage that BIP or linear
programming would allow us to incorporate constraints which are not
confined to be the elements of a matrix, as is the case with the
Hungarian algorithm. One such use case is demonstrated in
\cite{park2014minimum}. However, in the present scenario, Hungarian
algorithm is found to be computationally more efficient compared to
BIP in obtaining same solution.

Even though, it is an initial work with a lot of scope for
improvement, we believe that the material presented in this paper
would provide a lot of useful insights which can be appreciated by the
readers.  The rest of this paper is organized as follows. The proposed
method is provided in Section \ref{sec:meth}. The analysis of
experimental results is provided in Section \ref{sec:expt} followed by
conclusion in Section \ref{sec:conc}.

%
%
%The proposed method is explained next in the paper.

\section{Proposed Approach} \label{sec:meth}

\begin{figure}[htbp]
  \centering
  \includegraphics[scale=0.3]{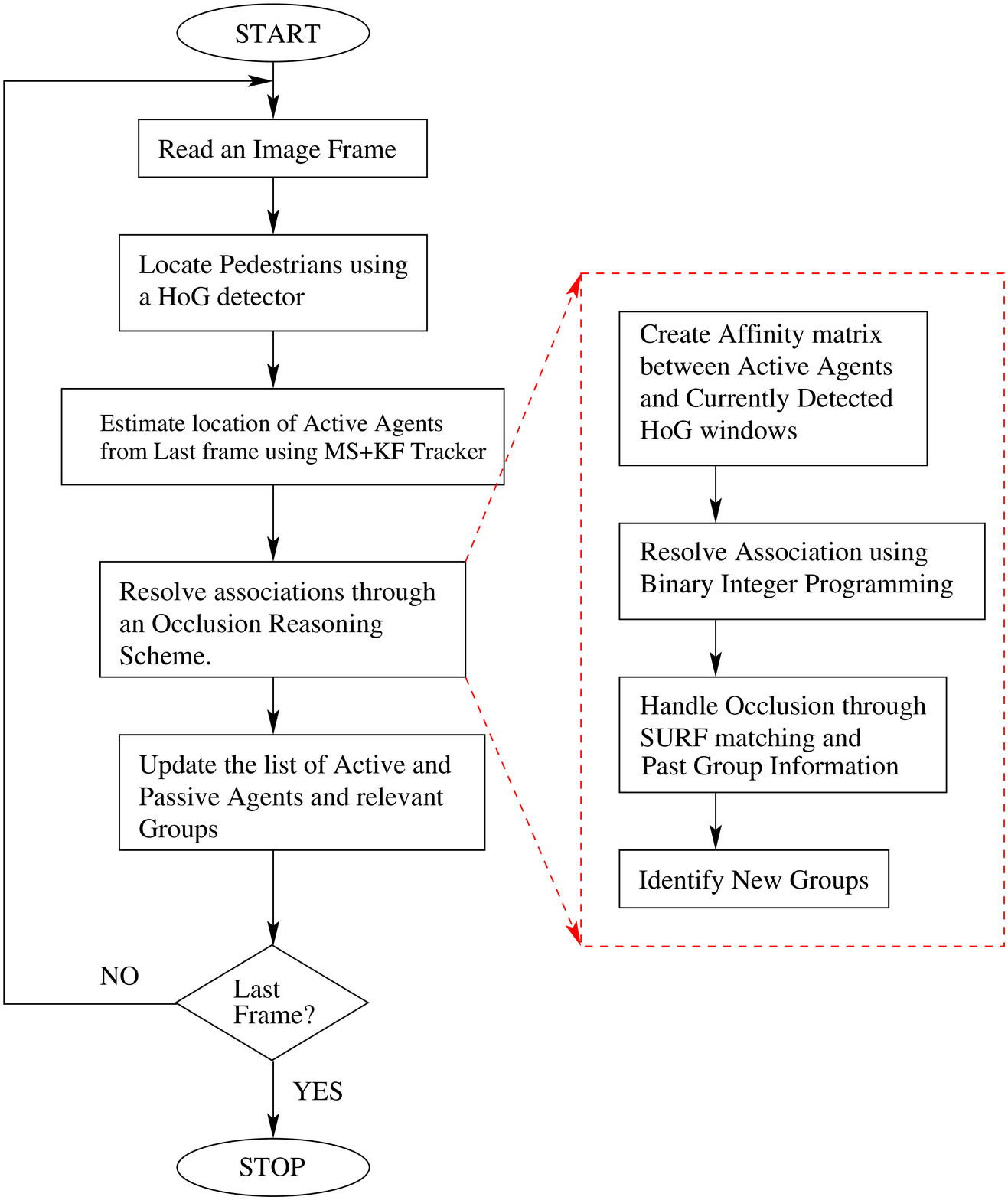}
  \caption{Outline of our approach for pedestrian tracking. The
  proposed occlusion reasoning scheme is shown in the red box. }
  \label{fig:fc}
\end{figure}

In order to explain our approach, we would use the following
notations. A given video sequence is represented by the symbol $I_k,\
k = 1,2,\cdots,N$ indicating that the video has a total of $N$ frames.
As stated earlier, any standard human detector is used to locate pedestrians
in each frame.  Each person detected or tracked is called an agent
which is represented by a bounding box (BB) surrounding the person and
is labeled with a global ID. The agents for a given frame $I_k$ is
represented by the symbol $A_k^i,\ i =1,2,\cdots,n$, where $n$ is the
number of agents that are found by the detector in the frame. 

The proposed method for carrying out pedestrian detection and tracking
is shown as a flowchart in Figure \ref{fig:fc}. The method consists of
four steps. The first step involves applying a human detector to
locate pedestrians in each frame. The second step involves estimating
the location of agents from the last frame using a tracker that uses
mean-shift algorithm           \cite{Comaniciu:2003} and a Kalman
Filter \cite{Haykin:2009}.  It is necessary to associate these agents
from the last frame with those obtained from the detector in the
current frame. The association simply means assigning appropriate
labels or IDs to the currently detected target windows. The problem
becomes difficult when the agents come together to form groups or
undergo partial or full occlusion.  We propose an \emph{occlusion
reasoning scheme} to solve this association problem between the past
agents and the currently detected target windows. This reasoning
scheme is explained next in this section. Once the associations are
resolved, the list agents is updated by adding new agents which are
found in the current frame. A list for the pairs of agents which
overlap with each other is also maintained which is also essential for
dealing with the cases of occlusion.  

\subsection{Occlusion reasoning scheme} \label{sec:ors}

The occlusion reasoning scheme includes four main steps.  The first
step involves creating an \emph{affinity matrix} between the estimated
agent windows obtained from the last frame (using tracker) and the
persons detected in the current frame by the detector. This matrix
is utilized in the next step to resolve association between these two
groups of agents using binary integer programming. The resulting
associations may contain few errors arising out of difficult cases
like occlusion.  Hence a second stage of verification based on SURF
matching and pairing information from the last frame.  Once the
associations with the previous agents are resolved, the newly detected
agents are given new agent IDs. The details for each of these steps
are provided below. 

\subsubsection{Affinity Matrix}\label{sec:affmat}

\begin{figure}[!h]
  \centering
  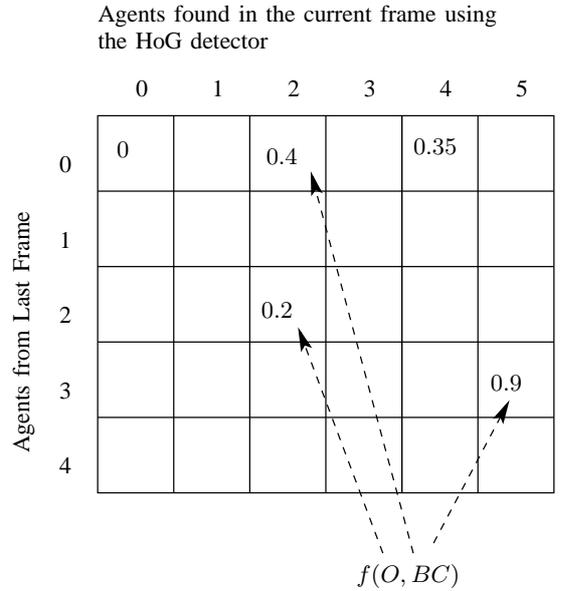
  \caption{Affinity Matrix for a given frame. A non-zero value
  indicates `closeness' between a pair of windows. The values are
normalized between 0 and 1.}
  \label{fig:affmat}
\end{figure}

An illustration of an affinity matrix for a given frame is shown in
Figure \ref{fig:affmat}.  Let us assume that the number of agents
found by the detector in the current frame $I_k$ is $n$ while the
number of agents obtained from the last frame ($I_{k-1}$) is $m$. The
location of these agents from the last frame is estimated using KF+MS
tracker. These estimated agents are represented by the symbol
$\hat{A}_{k-1}^k$. Hence the affinity matrix $S_k$ for
this frame has a dimension $m \times n$ with each element having a
value obtained from a scalar function given by:
\begin{equation} S_k(i,j) = f(O,BC) = \alpha_1 O(i,j) + \alpha_2
  BC(i,j) \label{eq:match-func} \end{equation} where $O(i,j)$ is the
percentage overlap between the two bounding boxes given by
\begin{equation} O(i,j) =   \frac{ \hat{A}_{k-1}^k(i) \cap A_k(j)
}{\hat{A}_{k-1}^k(i) \cup A_k(j)} \label{eq:overlap} \end{equation}
and $BC(i,j)$ is the Bhattacharya Coefficient computed between the
corresponding bounding boxes representing similarity based on
histogram matching. The weights $\alpha_1$ and $\alpha_2$ are
normalized weights which are decided a priori indicating the relative
importance of individual factors in the overall function. The values
of the matrix elements $S_k(i,j)$ lie between 0 and 1, 0 being no
overlap or similarity and 1 indicating high level of affinity or
similarity between the windows. This affinity matrix indicates the
`closeness' between a pair of windows. This matrix is used in the next
step to resolve associations between the agents obtained from the
previous frame and the persons detected in the current frame. 

\subsubsection{Resolving associations using Binary Integer Programming (BIP)} \label{sec:bip}

The association of currently detected target windows with those obtained
from the last frame is not straight forward. This is due to the fact
that this association depends on multiple features. The association
obtained using one feature might conflict with that obtained using
another feature. Secondly, there might be cases of one-to-many or
many-to-one associations between the two sets of windows. The first
cause is alleviated to some extent by forming the affinity matrix
where multiple features or criteria are combined to form a unique
scalar function that indicates the similarity or affinity between a
pair of windows. The second problem is solved by posing it as an
optimization problem which is solved by using binary integer
programming \cite{nemhauser1988integer}. The elements of affinity
matrix are considered to be the decision variables and constraints are
put over the rows and columns of the matrix, so that many-to-one or
one-to-many associations do not occur.  We use the COIN-OR CBC
\cite{forrest2005cbc} library in order to solve this problem. The
parameters of the proposed BIP formulation are as follows:     

\begin{itemize}
  \item $S_k(i,j) \in \mathbb{R}:$ coefficient of matching or
    similarity between a given pair of windows in the affinity matrix.

%  \item $t_k(i,j) \in \{0,1\}:$ binary coefficient corresponding to
%    $S_k(i,j)$, such that
%    \begin{equation}
%      t_k(i,j) = 
%      \begin{cases}
%        1, & \text{if } S_k(i,j)>0\\
%        0, & \text{otherwise}
%      \end{cases}
%    \end{equation}
  \item $u_{i,j} \in \{0,1\}:$ decision variable, $u_{i,j} = 1$, if
    the matching pair $(i,j)$ is selected, else $u_{i,j} = 0$.
\end{itemize}

The optimization problem is now stated as follows:

\begin{subequations} \label{eq:BIP}
  \begin{align}
    \arg \max_{u} \hspace{0.5cm} & \sum\limits_{i=1}^m
    \sum\limits_{j=1}^n S_k(i,j) u_{i,j} & \label{eq:obj} \\
    \text{subject to } \hspace{0.5cm} & \sum\limits_{i=1}^m u_{i,j}
    \leq 1 & \forall j \label{eq:constr1} \\
    & \sum\limits_{j=1}^n u_{i,j} \leq 1 & \forall i
    \label{eq:constr2} \\
    & u_{i,j} \in \{0,1\} & \forall\ i,j \label{eq:bound}
  \end{align}
\end{subequations}

\begin{figure}[!t]
  \centering
  \includegraphics[scale=1.0]{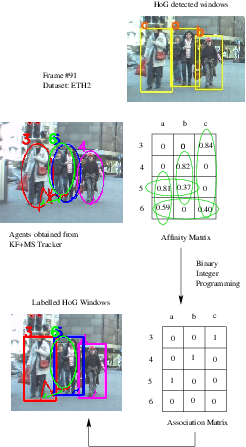}
  \caption{Resolving associations using Binary Integer Programming.
  The HoG detected windows (a,b,c) are associated with the agents
obtained from the tracker (3,4,5,6). Conflicting associations are
shown as green ellipse in the affinity matrix. The output of BIP is a
binary association matrix providing final labels for the HoG windows.
Agent 6 is not associated with any HoG window.}
  \label{fig:bip} 
\end{figure}

The objective function \eqref{eq:obj} aims at maximizing the
association between a pair of windows as given by the affinity matrix.
The constraints \eqref{eq:constr1} and \eqref{eq:constr2} allow only
one-to-one association between the considered pair of windows. The
bound \eqref{eq:bound} restricts the decision variable to be binary.
The decision variables having value of $1$ in the BIP solution
correspond to the selected pair of windows. The optimization process
for resolving association is illustrated in Figure \ref{fig:bip}. We
consider frame number 91 in the ETH2 dataset. In this image, three
detected target windows are labelled as $a$, $b$ and $c$ respectively.
From the previous frame, 4 windows are obtained using KF+MS tracker.
These windows have labels 3,4,5 and 6. The conflicting associations
arise due to the pairs shown in green ellipse in the affinity matrix.
The BIP module gives rise to a binary matrix providing unique
associations between the two sets of windows. Hence, the target window
$a$ is assigned the label 5, $b$ is assigned 4 and $c$ is assigned
label 3. The window ID $6$ is not associated with any target window
and hence appears as an ellipse in the final image.

\subsubsection{Occlusion Handling through SURF Matching} \label{sec:oh}

We will use some additional notations in order to explain the method
presented in this section. As stated above, each new agent is assigned
an unique global ID which is used to identify this agent wherever it
is visible in a video. We use the notation $L(A_k(i))$ to denote
the global ID of a given agent in the frame $I_k$. The symbol
$\{L(A_k)\}$ refers to the set of labels for all active agents
in this frame. We also define a set $G_k$ that consists of all 
pairs of agent IDs that overlap with each other. In other words, 
\begin{equation}
  G_k  \triangleq  \{ g_k^i \},\; i=1,2,\cdots,r   
  \label{eq:group1}
\end{equation}
where each element $g_k^i$ is a pair of agent labels (IDs) given by
\begin{equation}
  g_k^i = \{L(A_k(p)), L(A_k(q))\}, p\ne q, (p,q) \in  \{0,1,\dots,n\}
  \label{eq:group2}
\end{equation}

As explained in the previous section, the global labels of the active
agents obtained from the detector in the current frame is resolved
by the binary integer programming (BIP) module which uniquely assigns
the labels of agents from the last frame $L(A_{k-1})$ to the
currently detected agents. Let us denote this set of labels for the
currently detected agents by the symbol $\{L^{-}(A_k)\}$. Some of the
labels obtained from the BIP module might be erroneous, particularly
for those agents which get occluded or appear in groups in the current
frame. This is due to the fact that the decision of the BIP module
solely depends on the features used in the affinity matrix. Even
though multiple features or cues will provide robustness, yet it can
guarantee correct decisions for all cases.

\begin{figure}[!h]
  \centering
  \includegraphics[scale=1.0]{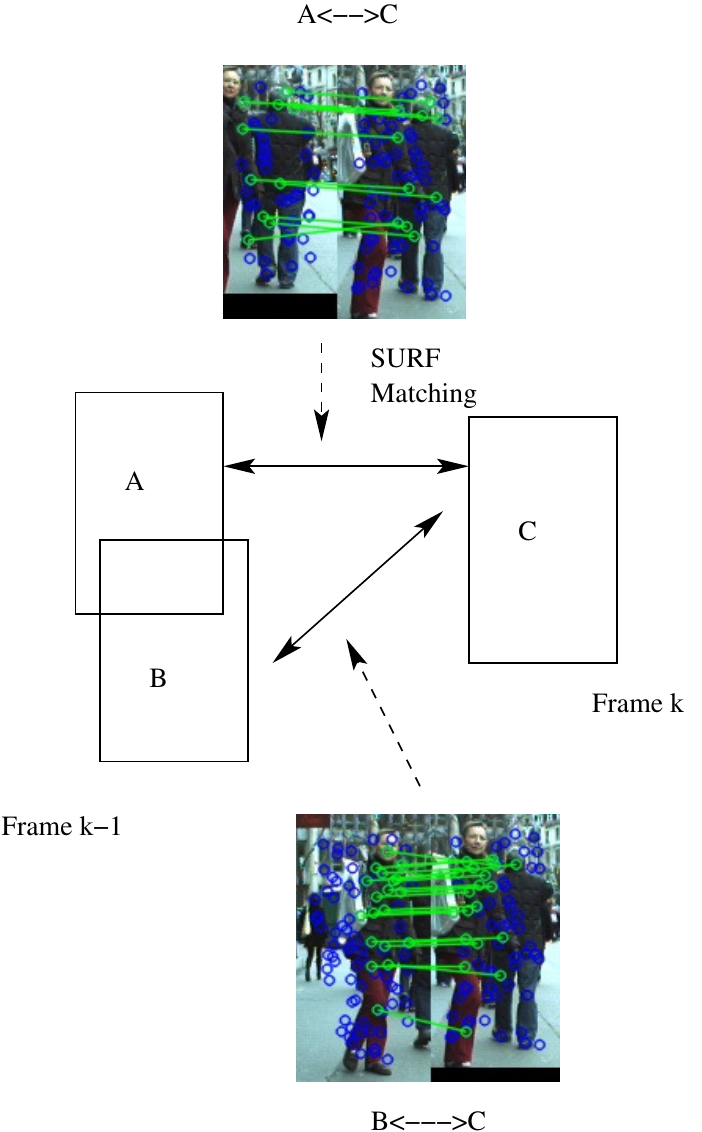}
  \caption{Occlusion Handling through SURF Matching. If one of the
  agents from the previous pairs is not present in the current frame,
use SURF-matching to identify the available agent.}
  \label{fig:surf-match}
\end{figure}

Therefore, a second stage of verification is employed to correct these
labels by using the pairing information obtained from the last frame
$G_{k-1}$ and SURF matching as explained in Algorithm \ref{alg:oh}. In
this algorithm, $n(G_{k-1})$ refers to the cardinality of the set
$G_{k-1}$. The basic idea is that if one of the agents in the pair
disappears in the current frame, SURF matching is used to recognize
the agent which is available, and is assigned the corresponding agent
label. The new set of global IDs obtained for the currently active
agents is denoted by $\{L(A_k)\}$.  Once the labels are found, a new
set of agent pairs are found based on whether they overlap or not.
This group is denoted by $G_k$ and will be utilized in the next
iteration.  The scheme is explained pictorially in Figure
\ref{fig:surf-match}. Let us assume that the agents $(A,B)$ form a
pair in the previous frame $I_{k-1}$ and in the current frame only one
window is detected by the detector. Let us call it $C$. Also
assume that the BIP module associates window $C$ with $A$. In this
case, the SURF matching between the pair (A,C) and (B,C) is used to
confirm the final  association. The one with maximum percentage match
is selected as the correct association pair.

\begin{algorithm}
  \caption{Occlusion Handling through SURF-Matching}
  \label{alg:oh}
  \begin{algorithmic}[1]
    \FOR{$i=0$ \TO $n(G_{k-1})$}
    \STATE $g_{k-1}^i = \{L(A_{k-1}(p)), L(A_{k-1}(q))\}$
    \IF{both the labels are in $\{L^{-}(A_k)\}$}
    \STATE Do nothing
    \ELSIF {both the labels are not present} %in $\{L^{-}(A_k)\}$}
    \STATE Do nothing
    %\ELSIF{ \{$L(A_{k-1}(p)) \in \{L^{-}(A_k)\}$\} \OR
    %\{$L(A_{k-1}(q)) \in \{L^{-}(A_k)\}$\}}
    \ELSE 
    [only one of the two labels, say, $p$ is present] %in  $\{L^{-}(A_k)\}$]
    \STATE Let $t$ be the index s.t. $L^{-}(A_k(t)) = L(A_{k-1}(p))$
    \STATE Compute SURF matching within the pairs $A_{k-1}(p) \sim
    A_k(t)$ and $A_{k-1}(q) \sim A_k(t)$ 
    %\STATE $A_k(t)$ is assigned the label of agent $A_{k-1}(s)$ for
    %which the SURF matching  ($A_{k-1}(s) \sim A_k(t)$, $s\in(p,q)$)
    %is maximum.
    \STATE New label to $A_k(t)$ is assigned as follows: 
    $ L(A_k(t)) = L(A_{k-1}(s))| s = \arg \max_s \{ A_{k-1}(s) \sim
  A_k(t), s \in(p,q)\}$
    \ENDIF
    \ENDFOR
  \end{algorithmic}
\end{algorithm}

\subsection{Estimating agent location using Kalman Filter and Mean-shift tracker}

As stated earlier, the location of agents from the last frame is
estimated in the current frame using a mean-shift tracker combined
with a Kalman Filter. It is well known that the detector may not
provide detection for a given agent in every frame where it is
located. In case of detection failure, a Kalman Filter could be used
to predict its location. The Kalman Filter itself learns from the
observations obtained from the  detector. The reliance on object
detector which is computationally expensive could be reduced by using
a mean-shift tracker \cite{Comaniciu:2003} that uses a colour
histogram to locate target in the next frame. The mean-shift tracker
is initialized for each new agent obtained from the detector. In cases
where the detector fails to locate this agent, the Kalman Filter and
the mean-shift tracker could be used together to confirm the location
of the said agent. Moreover, use of mean-shift tracker along with a
Kalman Filter could be used to reduce the computational cost by
reducing the search area for the detector.

\section{Experimental Results} \label{sec:expt}

\begin{figure}[!h]
  \centering
  \includegraphics[width=0.5\textwidth,height=7cm]{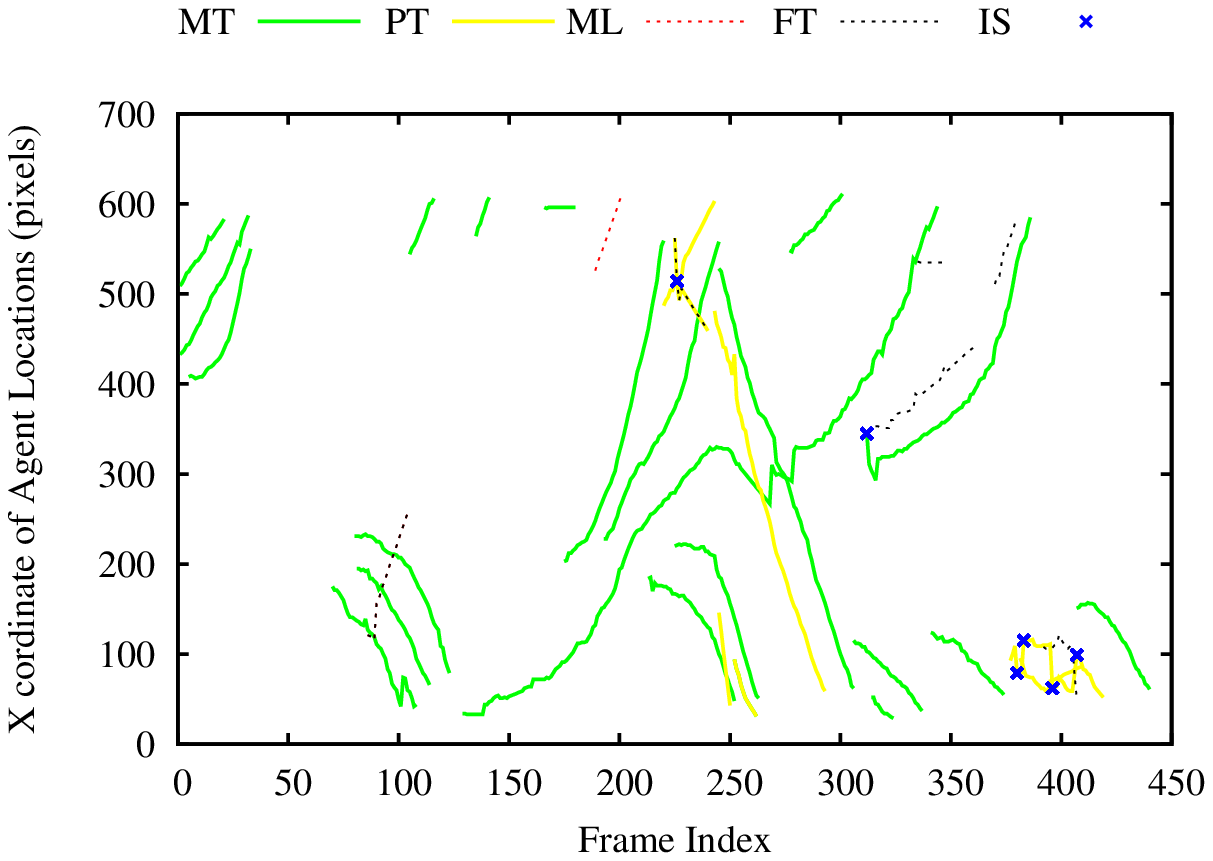}
  \caption{Tracking performance for ETH2 dataset. False trajectories
  are generated due to wrong detections by the HoG detector. Most
agents (shown in green) are correctly tracked with the help of the
proposed occlusion reasoning scheme.}
    \label{fig:traj}
\end{figure}

The performance of our algorithm is evaluated on three datasets which
are the same as used by Yang and Nevatia \cite{yang2014multi}: TUD
dataset \cite{andriyenko2011analytical}, PETS 2009 \cite{PETS:2009}
and ETH dataset \cite{ess2008mobile}. Since we wanted to compare our
results  with those reported in \cite{yang2014multi}, we used the same
detector and performance parameters to compute our tracking results.
The resulting comparison is provided in Table \ref{tab:perfcomp}. The
performance parameters used are: precision, recall, false alarm per
frame (FAF), ground truth (GT), mostly tracked trajectories (MT),
partially tracked trajectories (PT), mostly lost trajectories (ML),
number of trajectory fragmentation (Frag) and number of id switches
(IDS).  Please refer to the above paper for definitions of various
parameters mentioned in this table. Few such parameters computed for
one of the ETH datasets are shown in Figure \ref{fig:traj}. In this
figure, false trajectories (FT) are those which are generated due to a
false detection made by the HoG detector.  Some of the snapshots of
various agent trajectories for different datasets are shown in Figure
\ref{fig:snaps}. The complete tracking video is made available on web
\cite{pedtrack} for the convenience of readers. The snapshots show
some of the cases where our scheme is able to resolve associations
resulting in accurate tracking for the agent.

We can see that the performance of our algorithm is not good compared
to the Nevatia's latest work \cite{yang2014multi} even though we have
better tracking performance such MT, PT and ML. It is to be noted that
Nevatia's work is based on tracklets that introduces latency into the
decision making process unlike our approach where we take decision per
frame basis. However, this is an initial work which can be improved in
several ways. Some of them are as follows: 

1) We have more trajectory fragmentation and IDS because, we create
new IDs for the same agent if it remains occluded or not detected for
a certain number of frames. One approach would be to compare the
currently detected targets with not only with the last frame but also
past trajectories. 
2) We are using Kalman Filter as the motion predictor for each agent.
Probably, this assumption is not valid in case of camera motion. The
relative position of agents could be used as a parameter for resolving
associations between agents as suggested in \cite{yang2014multi}. 
3) The direction of each agent's motion along with motion coherence
can be incorporated into the affinity matrix. 
4) The values of $\alpha_i$ in equation \eqref{eq:match-func} is
decided a priori by the user. This could be treated as variables to be
optimized over another set of constraints. 
5) Taking cue from Nevatia's work \cite{yang2014multi}, the relative
location of pedestrians could be utilized to compensate for camera
motion.

\begin{table*}
  \centering
  \begin{tabular}{|c|c|c|c|c|c|c|c|c|c|} \hline
    Method & Recall (\%) & Precision (\%) & FAF & GT & MT (\%) & PT (\%) & ML
    (\%) & Frag & IDS \\ \hline\hline
    \multicolumn{10}{|c|}{ETH Dataset}\\ \hline \hline
    Our approach & 87.7 & 49.8 & 6.76& 125 & 78.2 & 17.8 & 4.0 & 182 & 27 \\ \hline
    Yang \& Nevatia (2014) &  79.0 & 90.4 & 0.637 & 125 & 68.0 & 24.8
    & 7.2 & 19  & 11 \\ \hline \hline
    \multicolumn{10}{|c|}{PETS 2009 Dataset}\\ \hline \hline
    Our approach & 97.3 & 68.0 &2.66 & 19 & 94.7 & 5.3 & 0.0 & 64 & 8 \\ \hline
    Yang \& Nevatia (2014) &  93.0 & 95.3 & 0.268 & 19 & 89.5 & 10.5
    & 0.0 & 13  & 0 \\ \hline \hline
    \multicolumn{10}{|c|}{TUD Dataset}\\ \hline \hline
    Our approach & 94.2 & 77.0 &1.74 & 10 & 100 & 0.0 & 0.0 & 10 & 3 \\ \hline
    Yang \& Nevatia (2014) &  87.0 & 96.7 & 0.184 & 10 & 70.0 & 30.0
    & 0.0 & 1  & 0 \\ \hline \hline
  \end{tabular}
  \caption{Comparison of Tracking results for different datasets}
  \label{tab:perfcomp}
\end{table*}

\begin{figure*}[!t]
  \centering
  \begin{tabular}{ccccc}
    \includegraphics[scale=0.22]{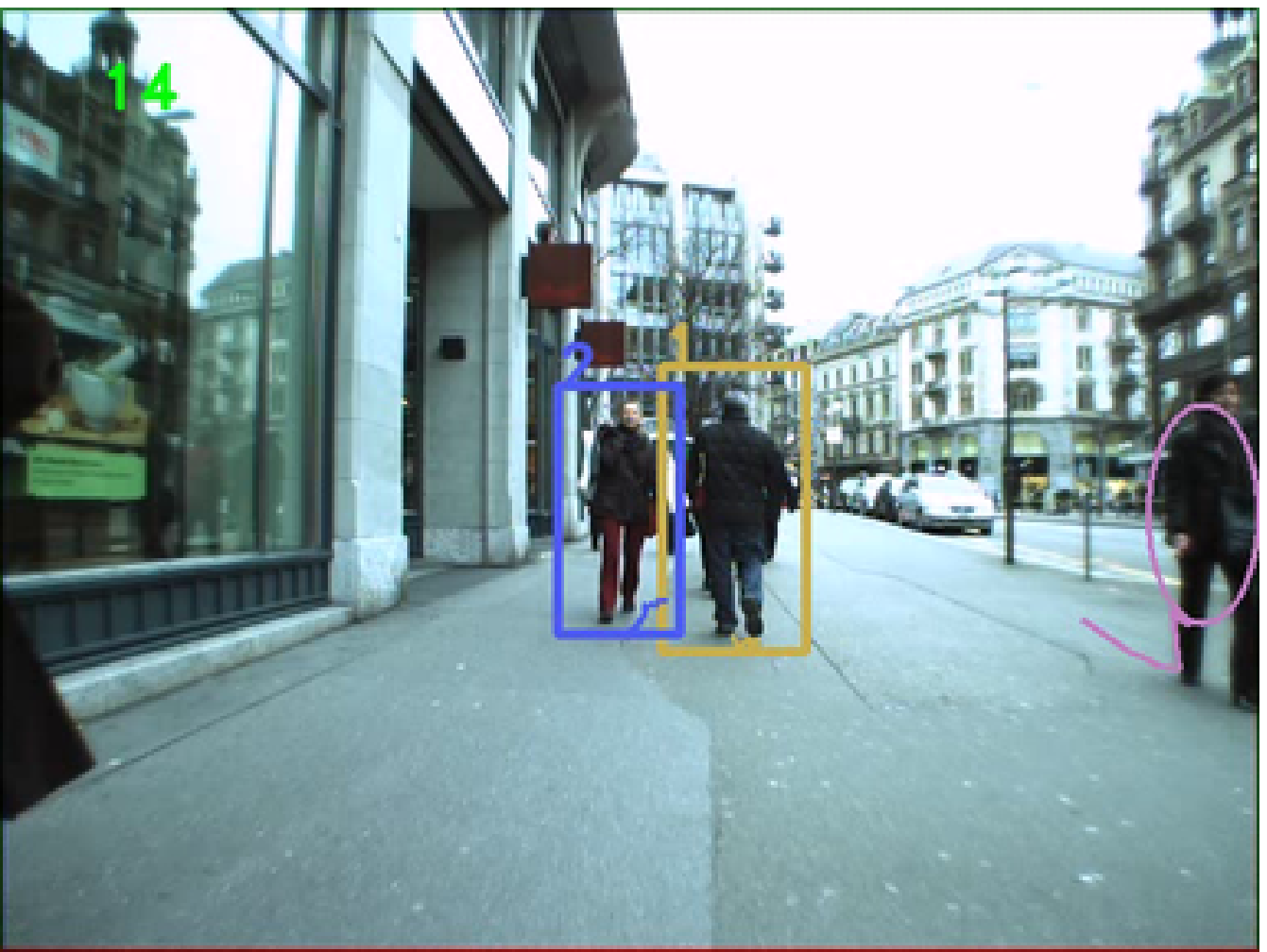} & 
    \includegraphics[scale=0.22]{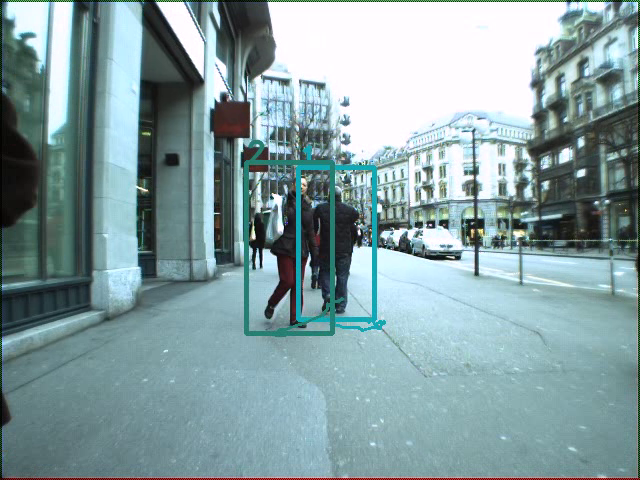} & 
    \includegraphics[scale=0.22]{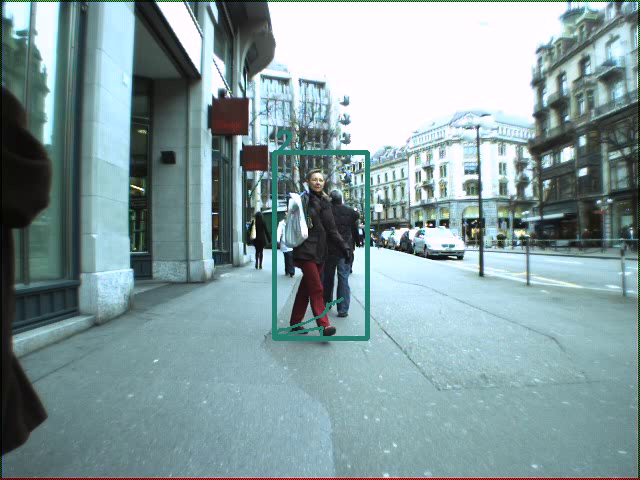} & 
    \includegraphics[scale=0.22]{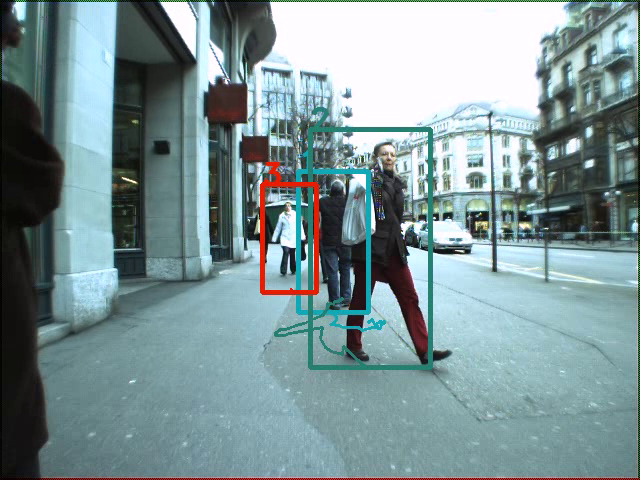} & 
    \includegraphics[scale=0.22]{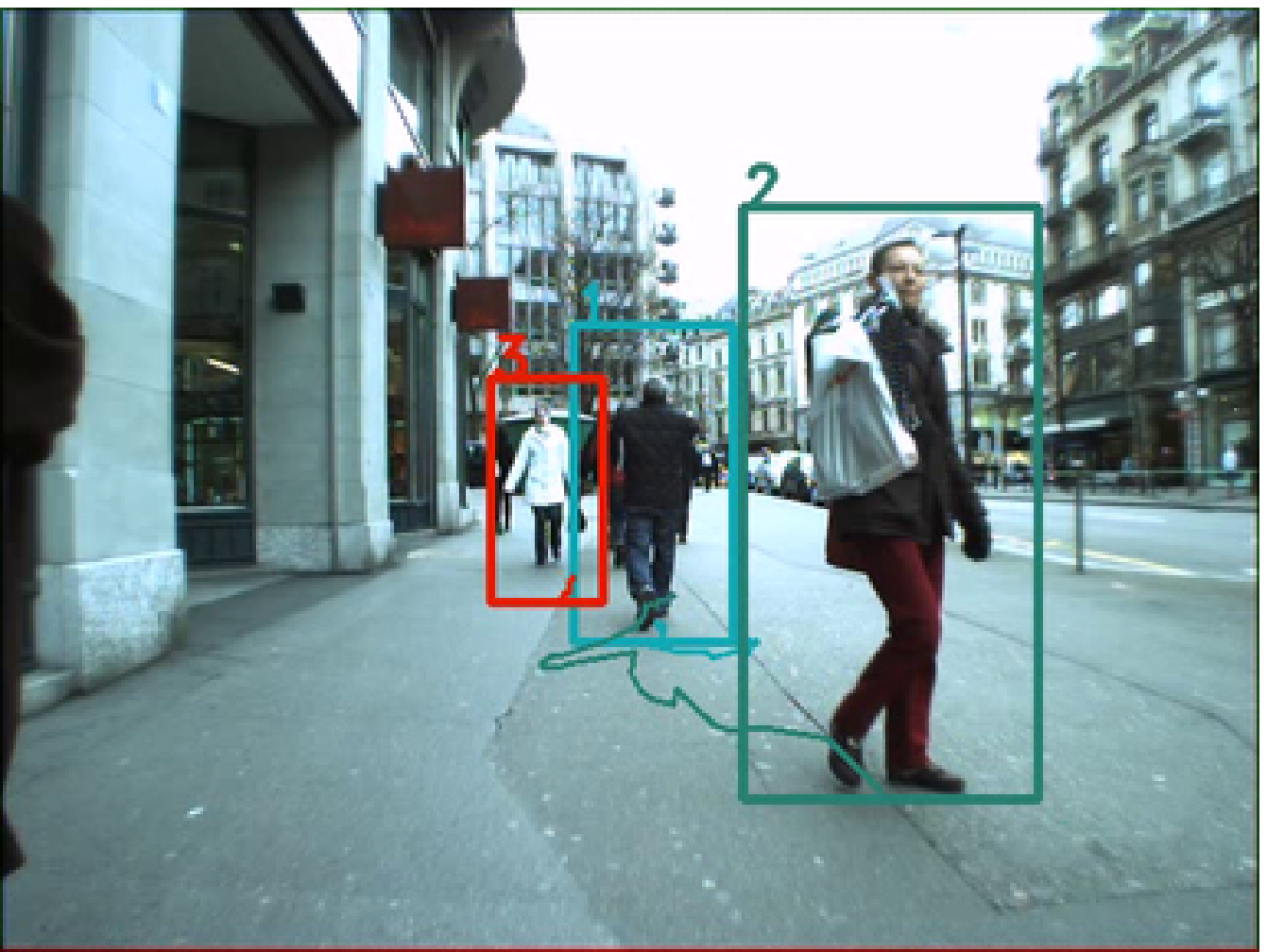} \\
    (1a) & (1b) & (1c) & (1d) & (1e) \\
    \includegraphics[scale=0.22]{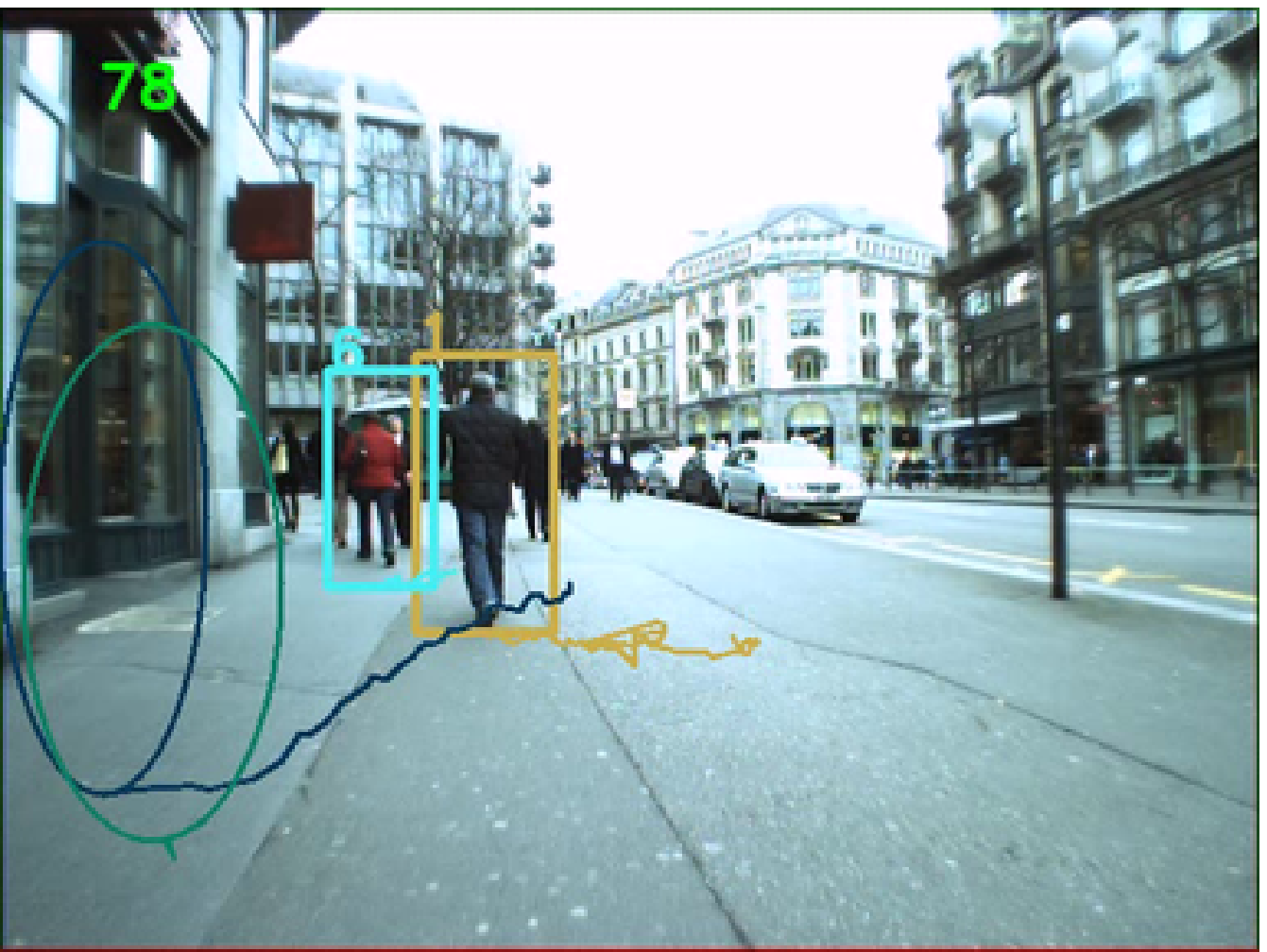} & 
    \includegraphics[scale=0.22]{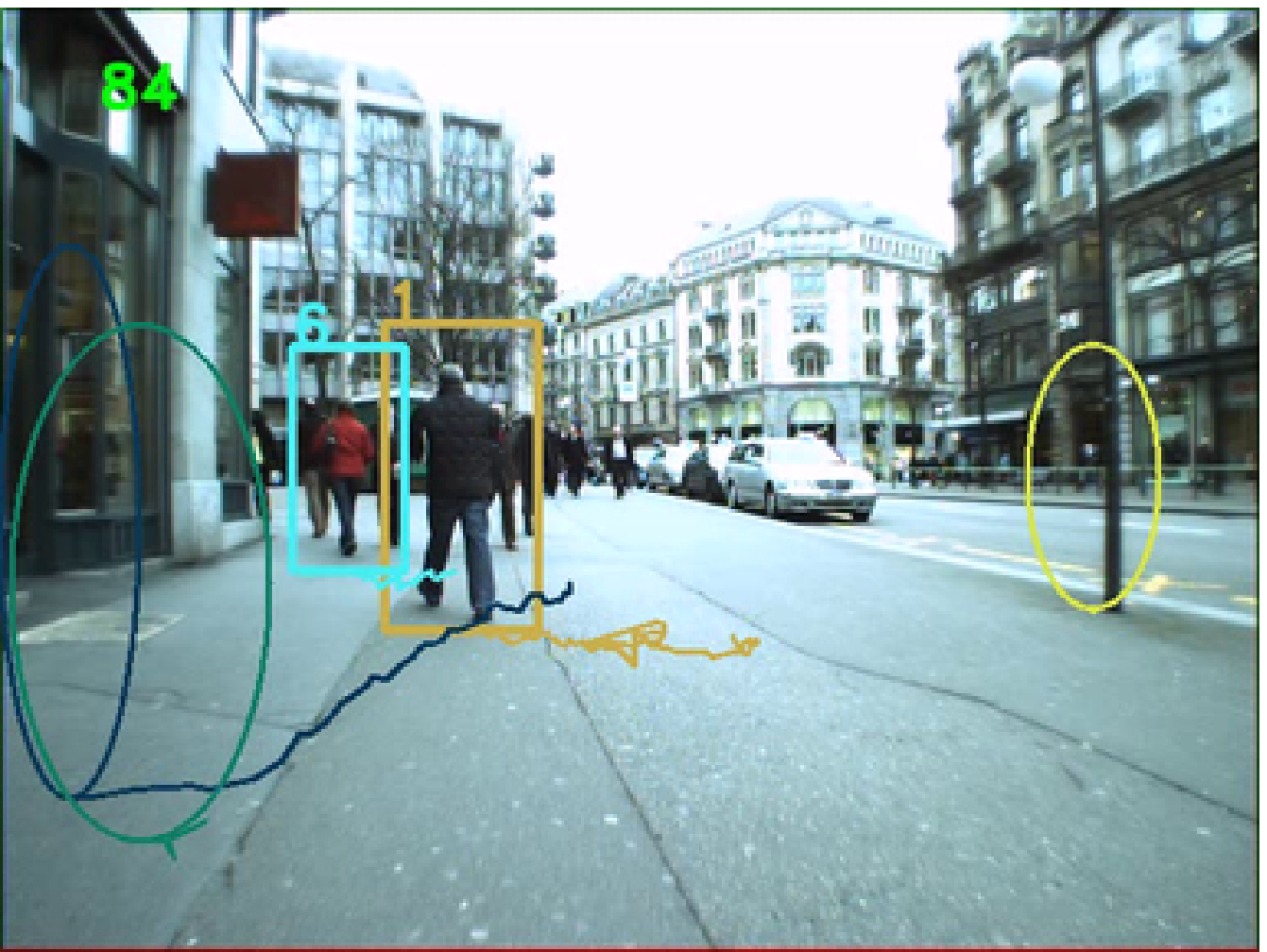} & 
    \includegraphics[scale=0.22]{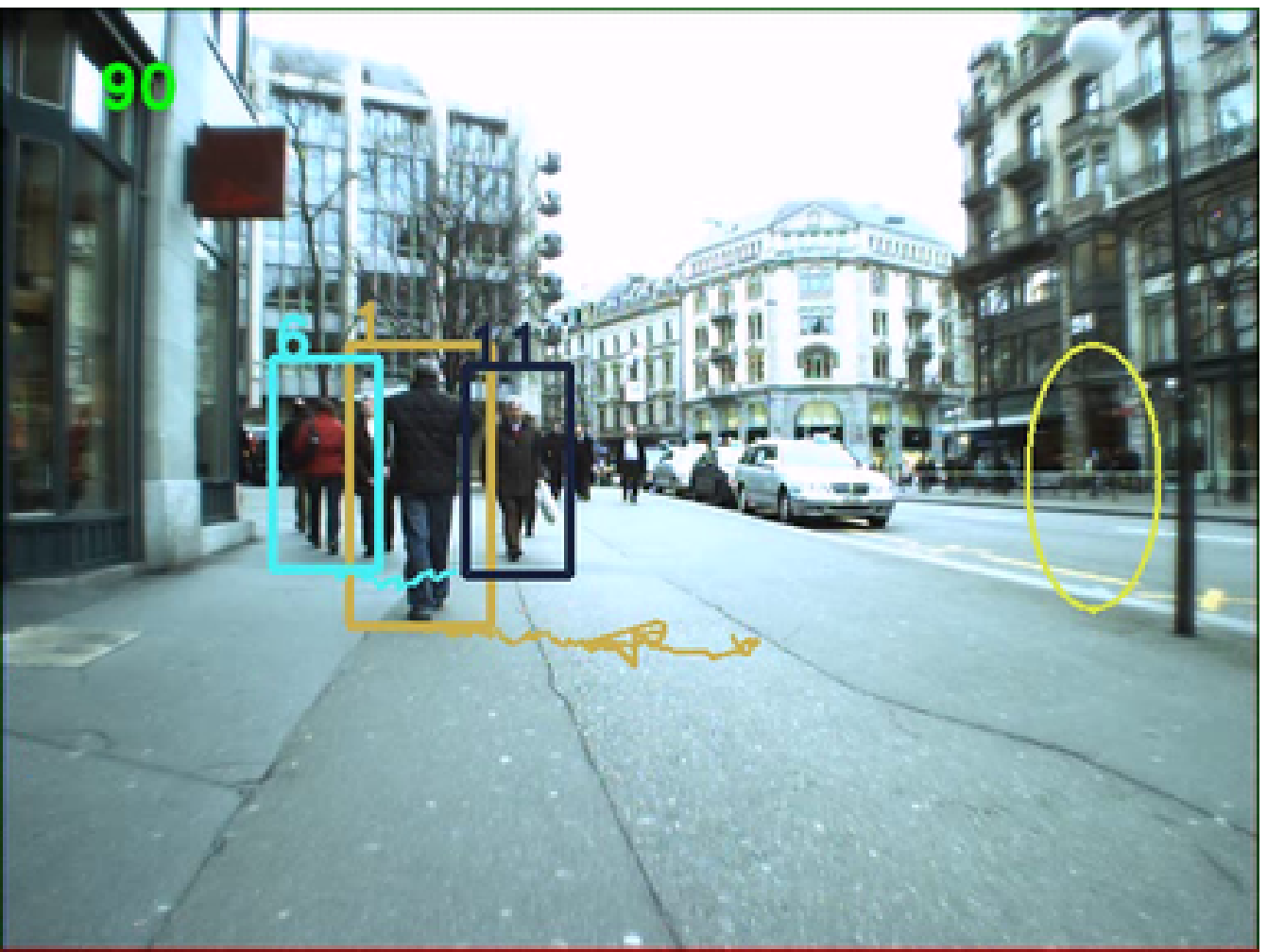} & 
    \includegraphics[scale=0.22]{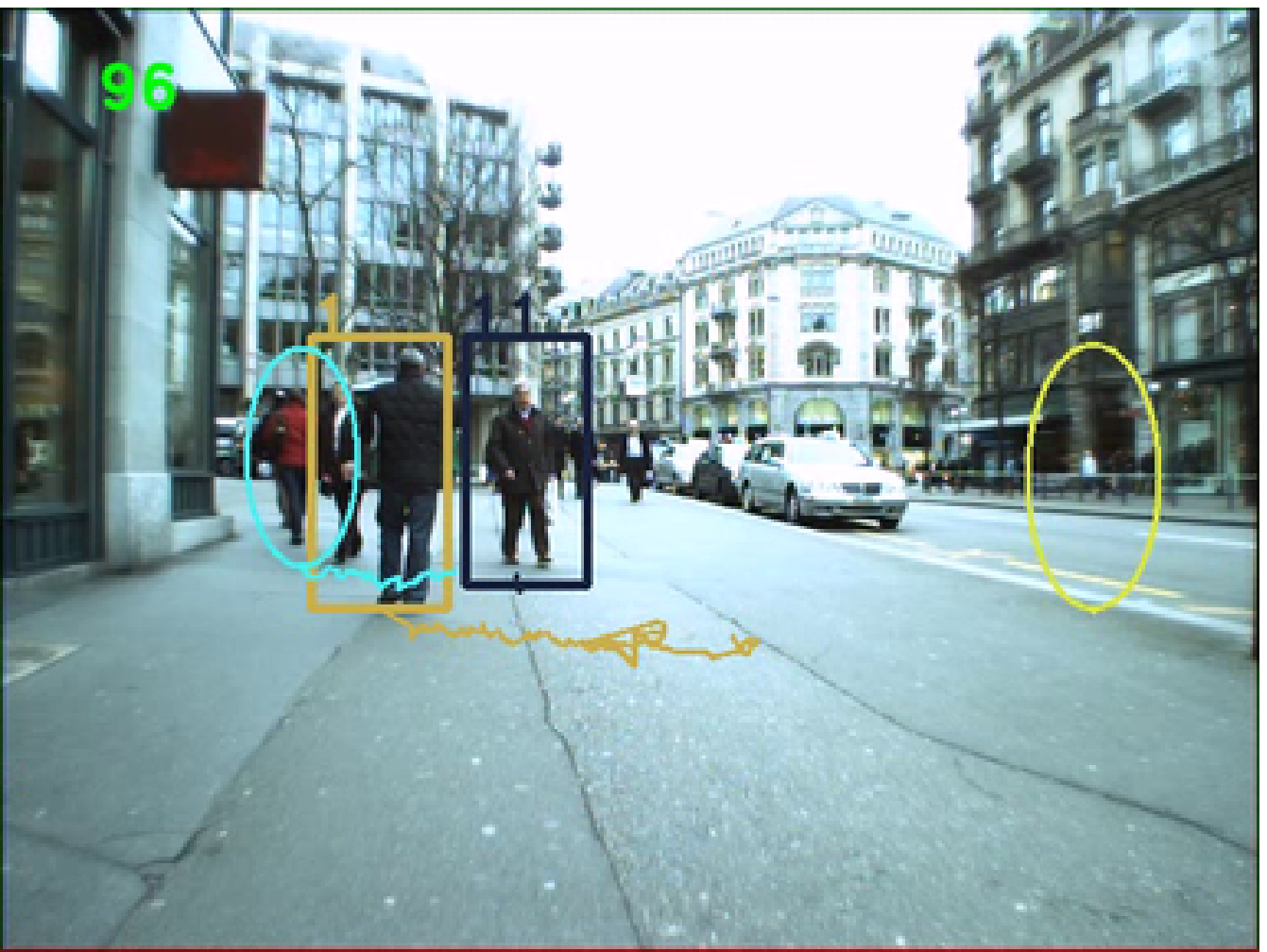} & 
    \includegraphics[scale=0.22]{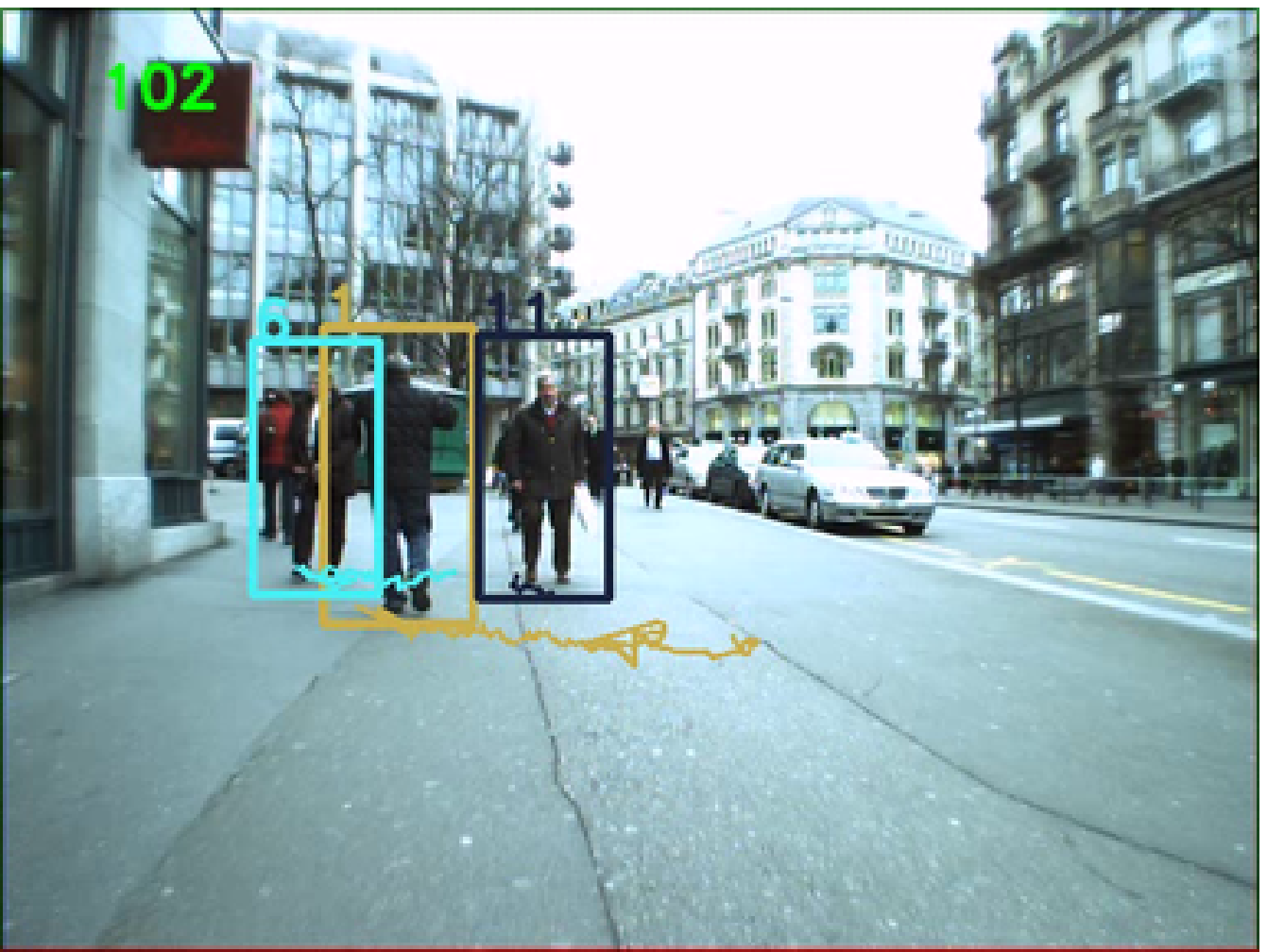} \\
    (1f) & (1g) & (1h) & (1i) & (1j) \\
    \multicolumn{5}{c}{\parbox{0.9\textwidth}{ETH1 dataset: (1a-1e)
      Trajectory of agent 2 (lady on left with blue window) intersects
      with that of agent 1 (person in the centre) without any ID
      switch. (1f-1j): Agent 1 (brown) is tracked successfully even
    when it forms group with other agents. However ID switch occurs
  with agent 6 (cyan). }}\\
    \includegraphics[scale=0.22]{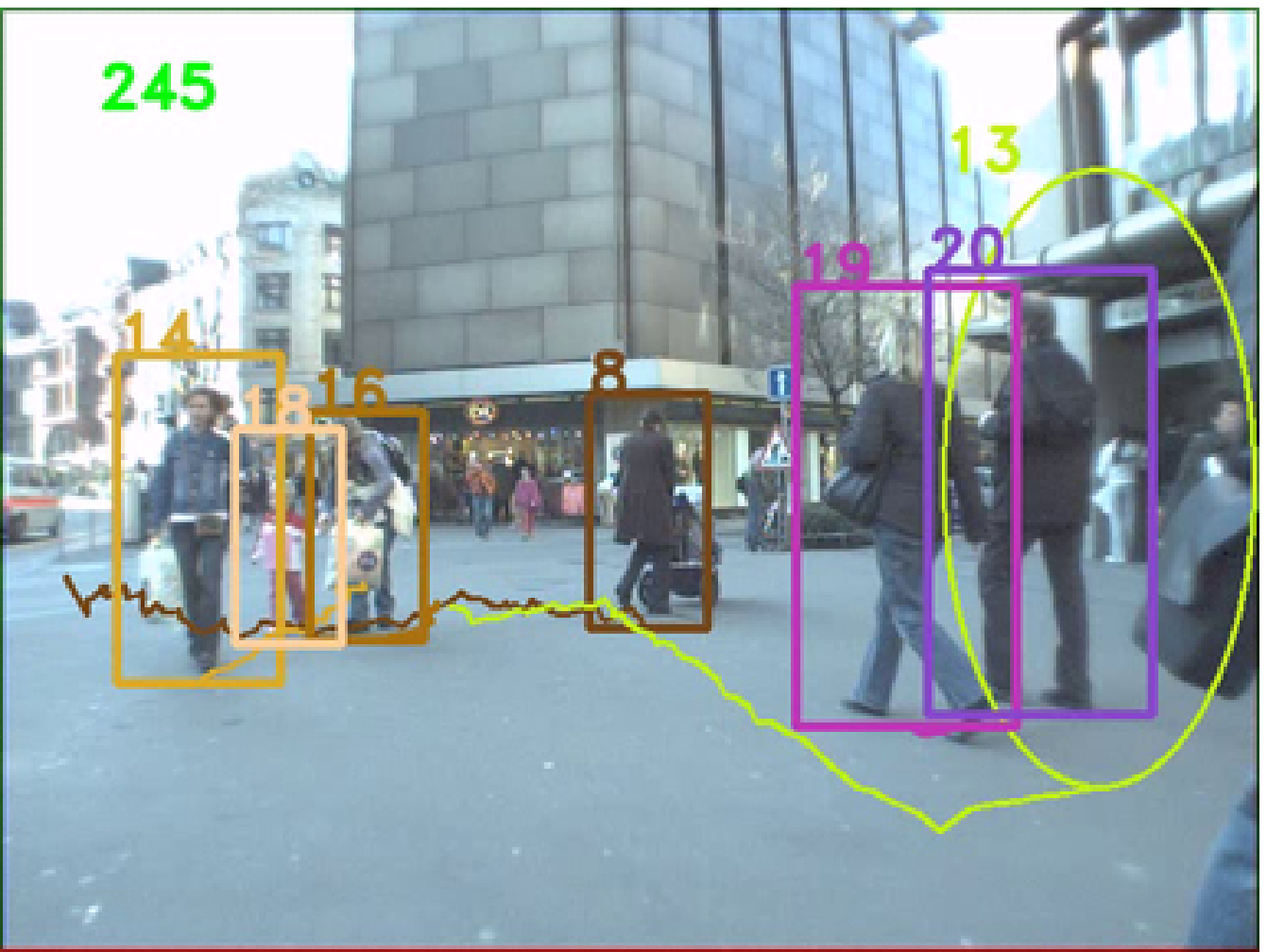} & 
    \includegraphics[scale=0.22]{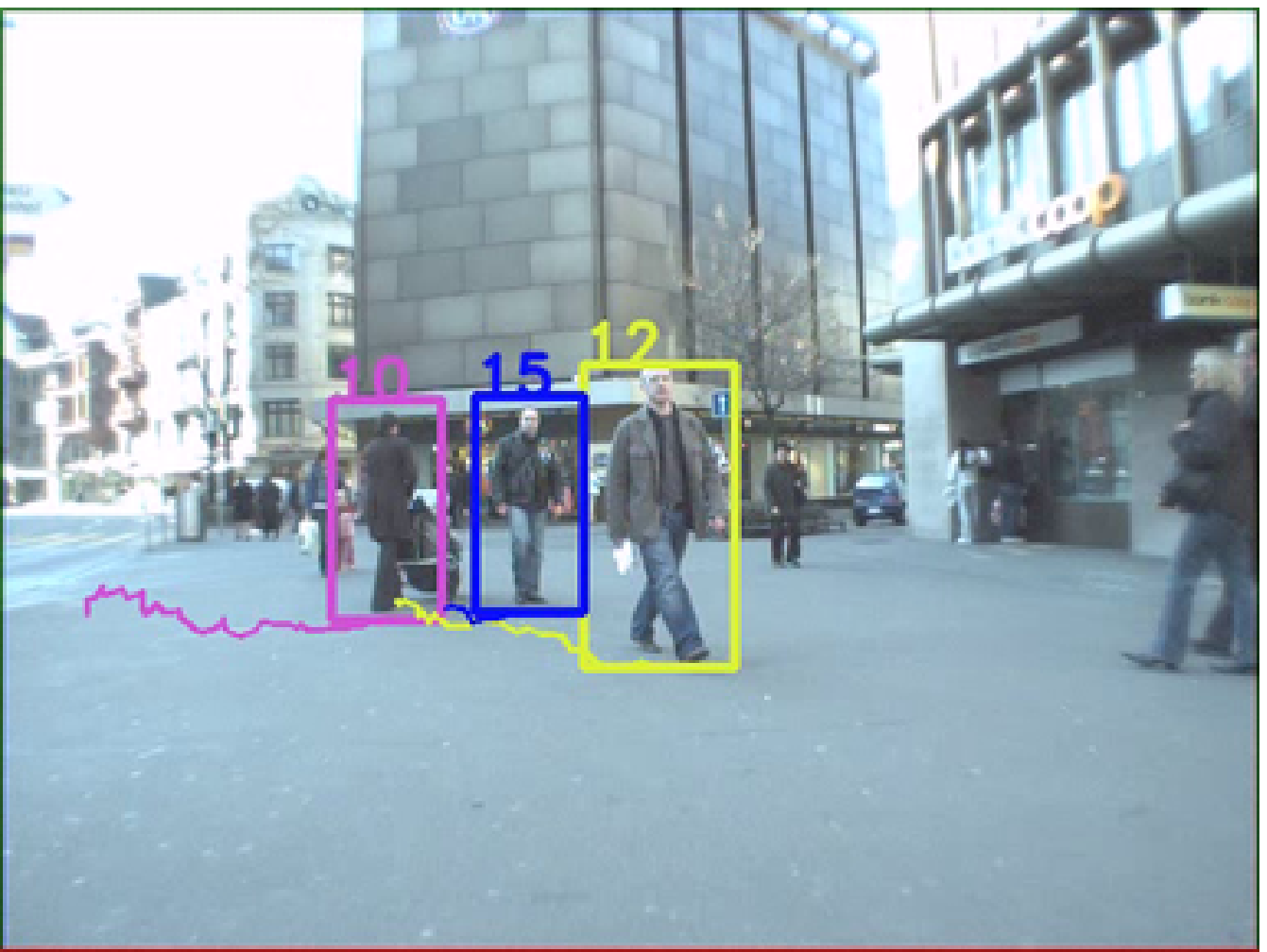} & 
    \includegraphics[scale=0.22]{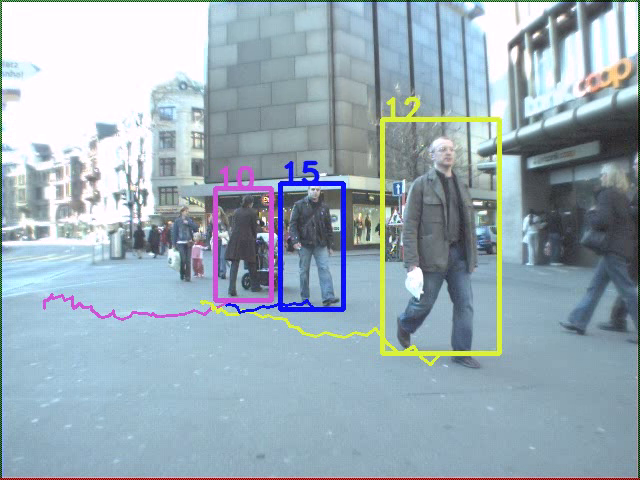} & 
    \includegraphics[scale=0.22]{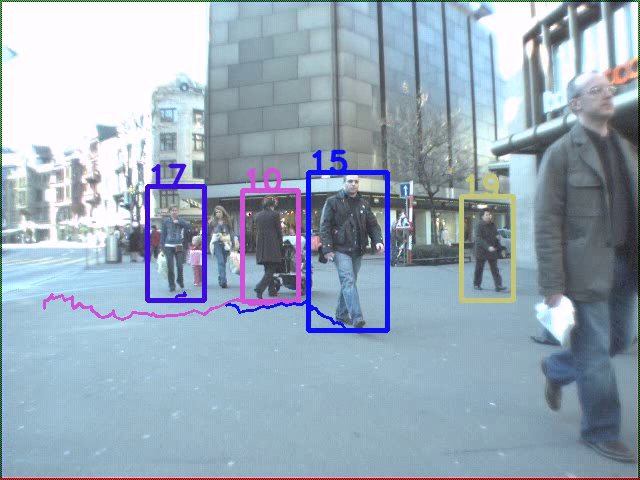} & 
    \includegraphics[scale=0.22]{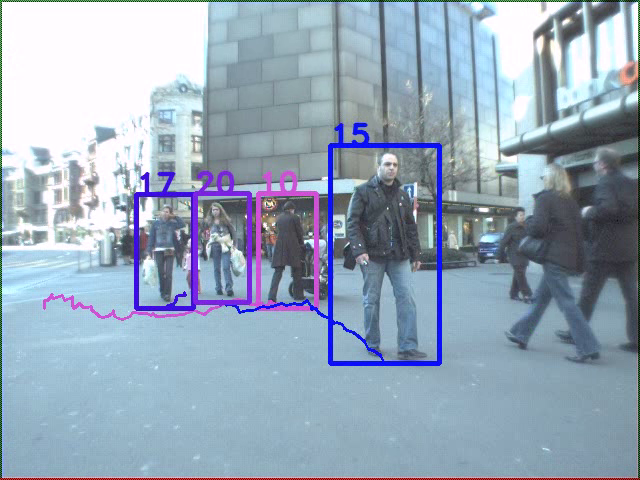} \\
    (2a) & (2b) & (2c) & (2d) & (2e) \\
    \includegraphics[scale=0.22]{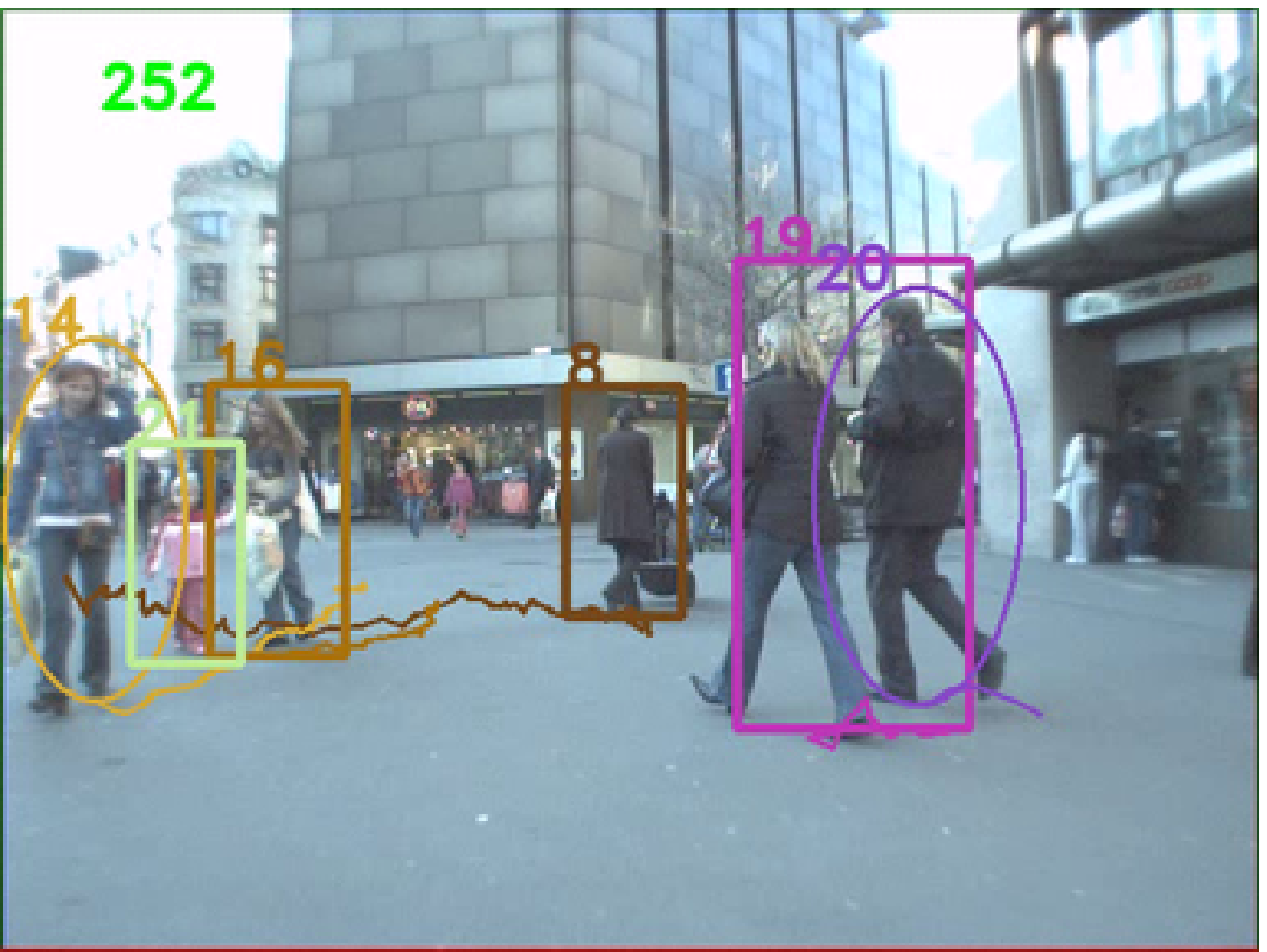} & 
    \includegraphics[scale=0.22]{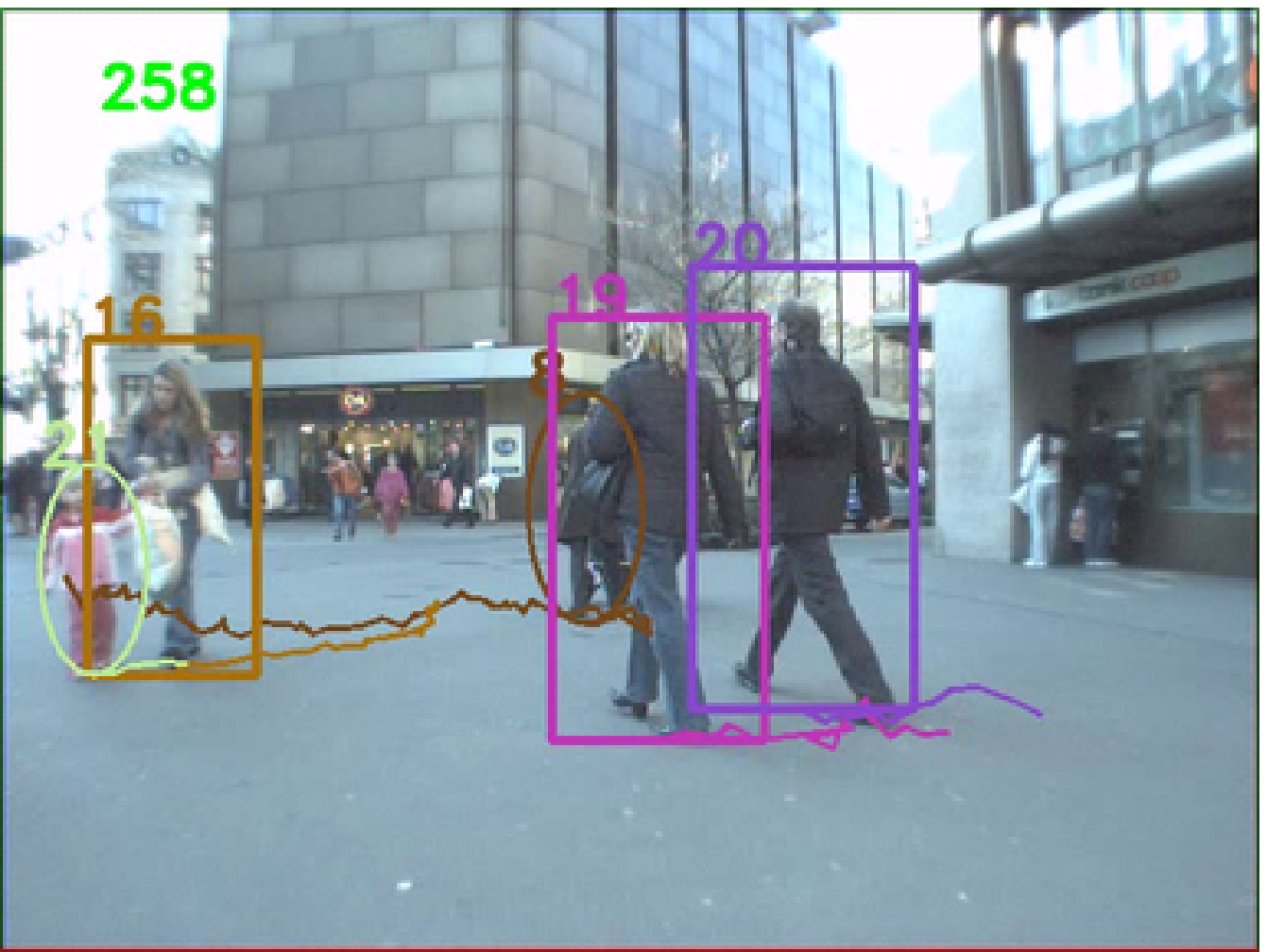} & 
    \includegraphics[scale=0.22]{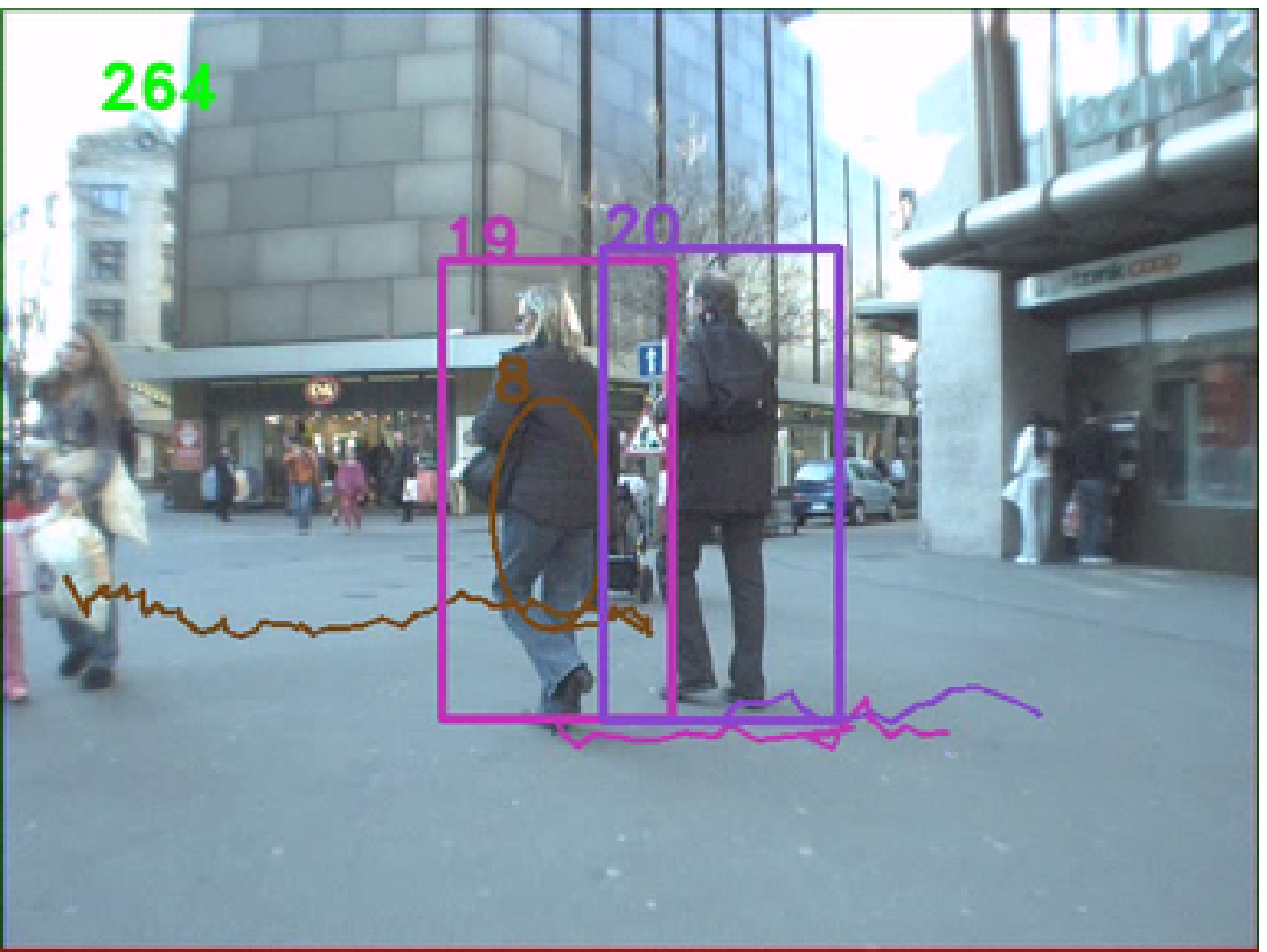} & 
    \includegraphics[scale=0.22]{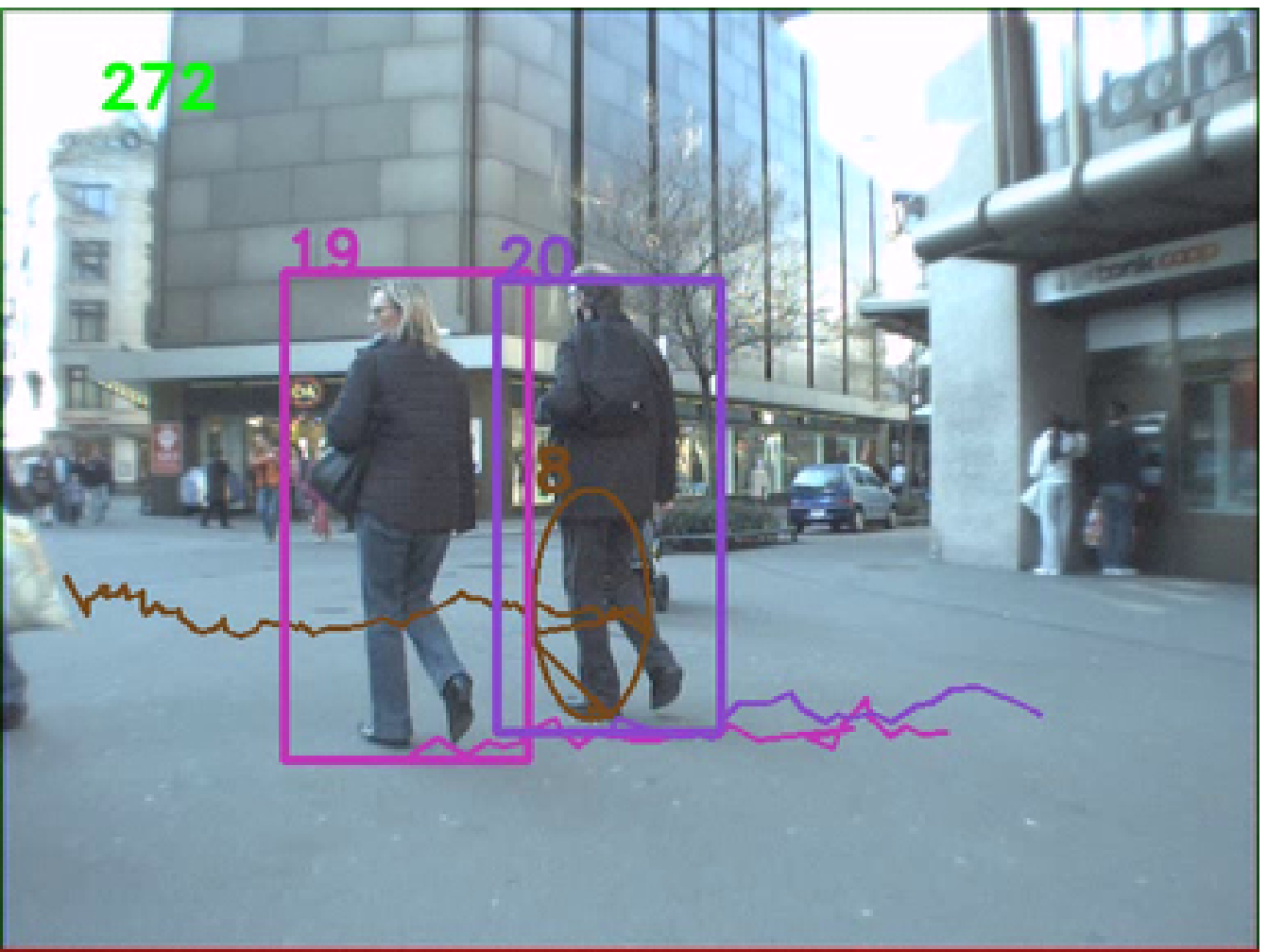} & 
    \includegraphics[scale=0.22]{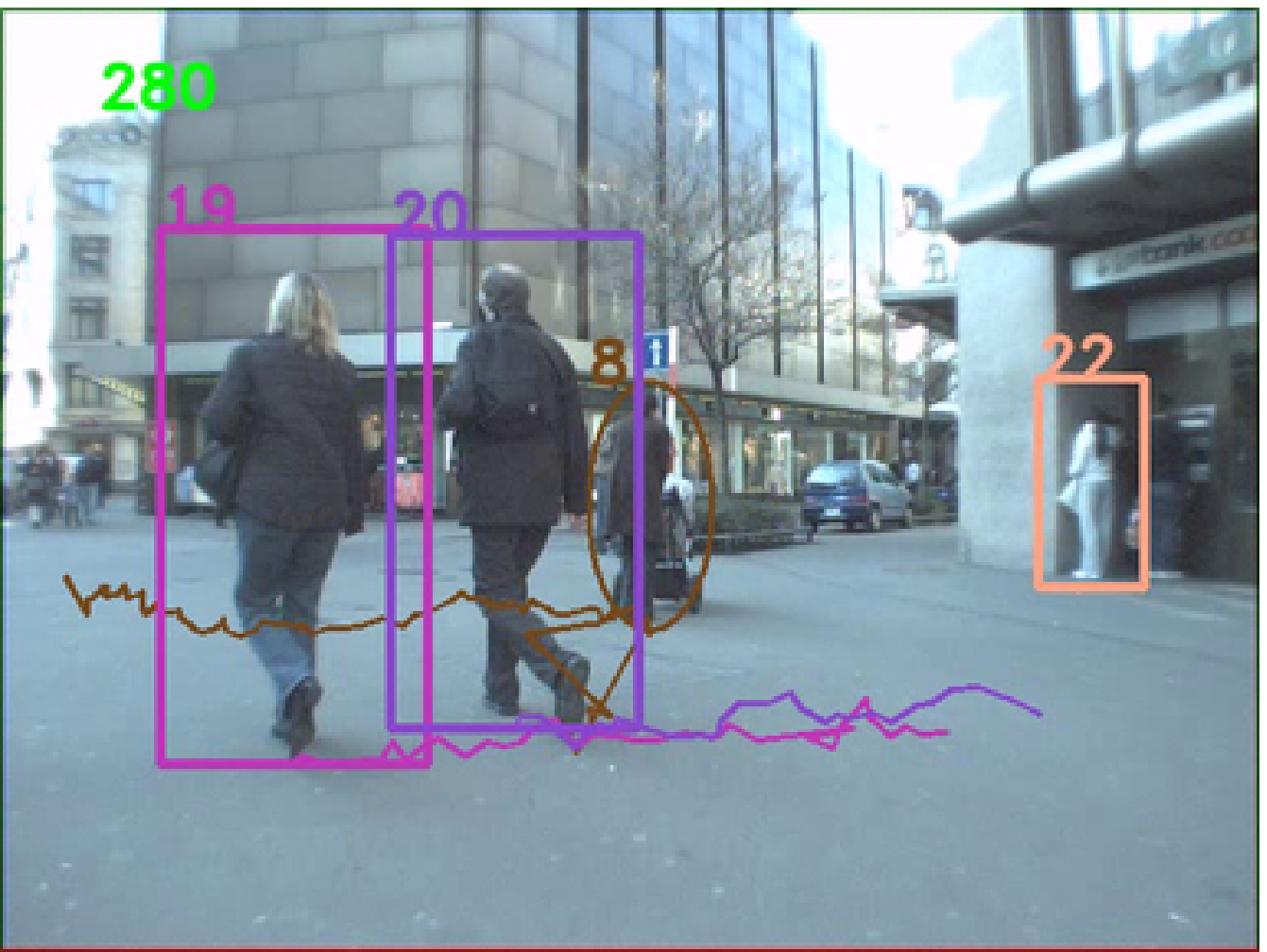} \\
    (2f) & (2g) & (2h) & (2i) & (2j) \\
    \multicolumn{5}{c}{\parbox{0.9\textwidth}{Dataset ETH2: (2a-2e)
    Agent 10 (lady in pink box) is tracked successfully through the
  crowd in spite of several instances of occlusion. (2f-2j) All the
agents are tracked successfully inspite of grouping and occlusion. }}\\
    \includegraphics[scale=0.22]{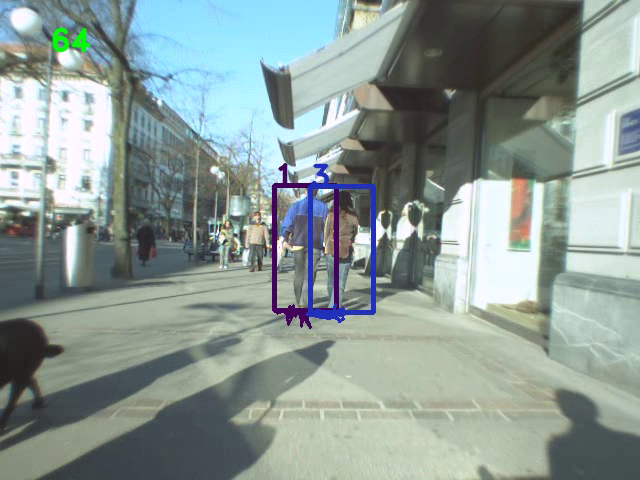} & 
    \includegraphics[scale=0.22]{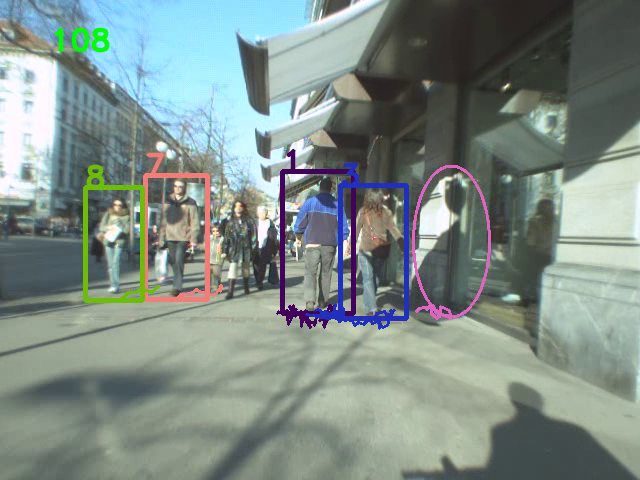} & 
    \includegraphics[scale=0.22]{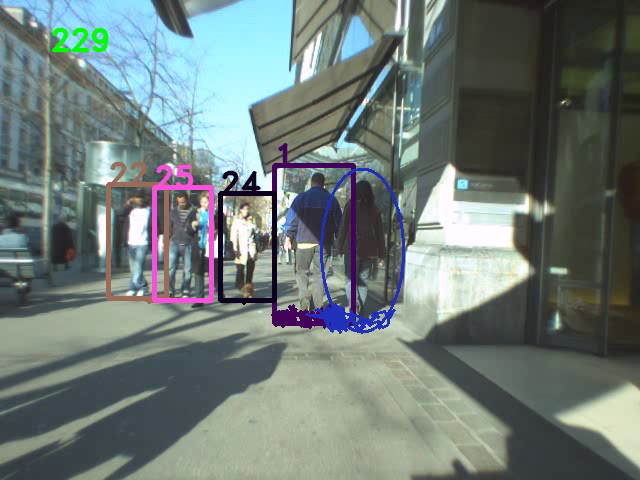} & 
    \includegraphics[scale=0.22]{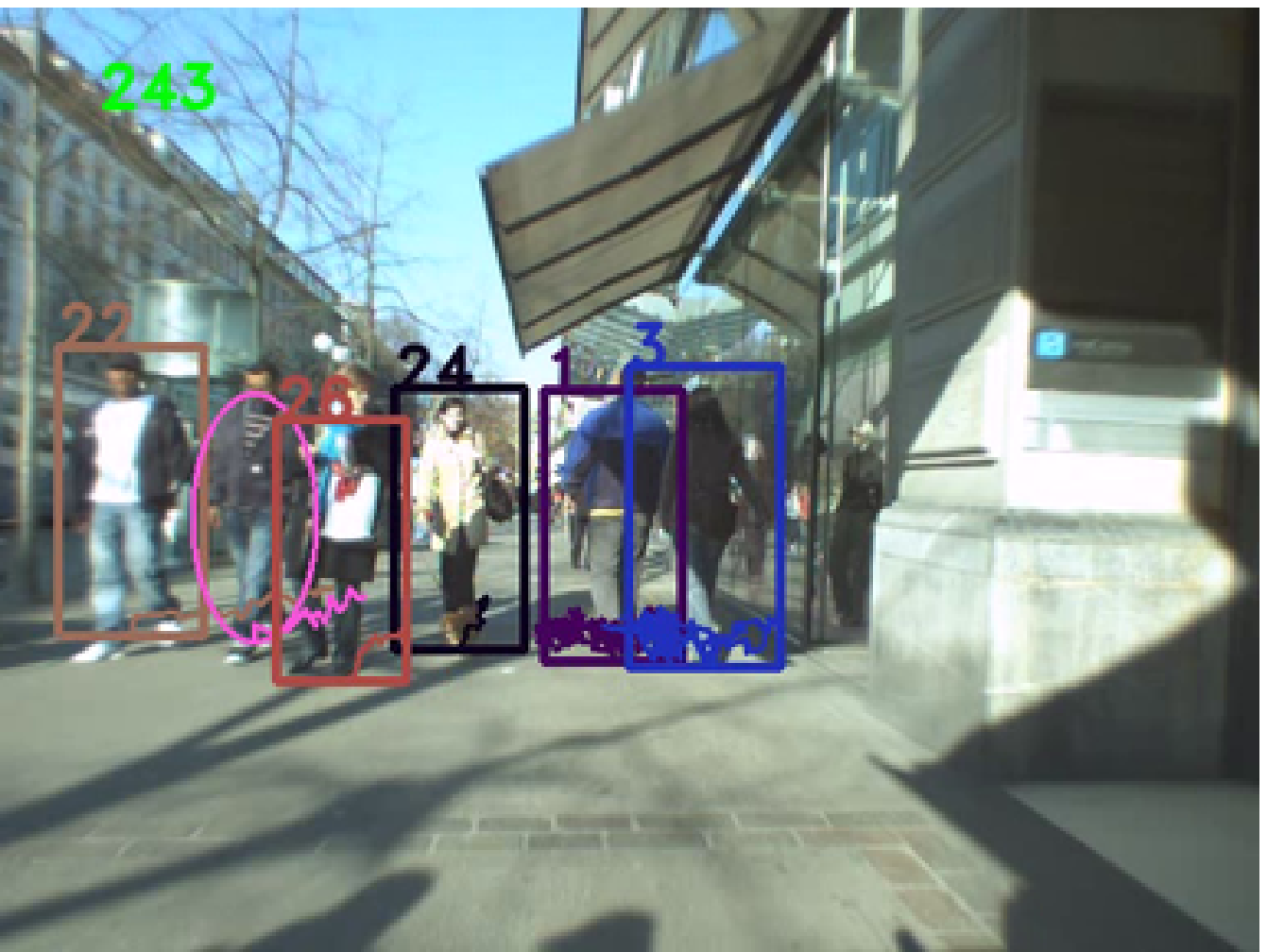} & 
    \includegraphics[scale=0.22]{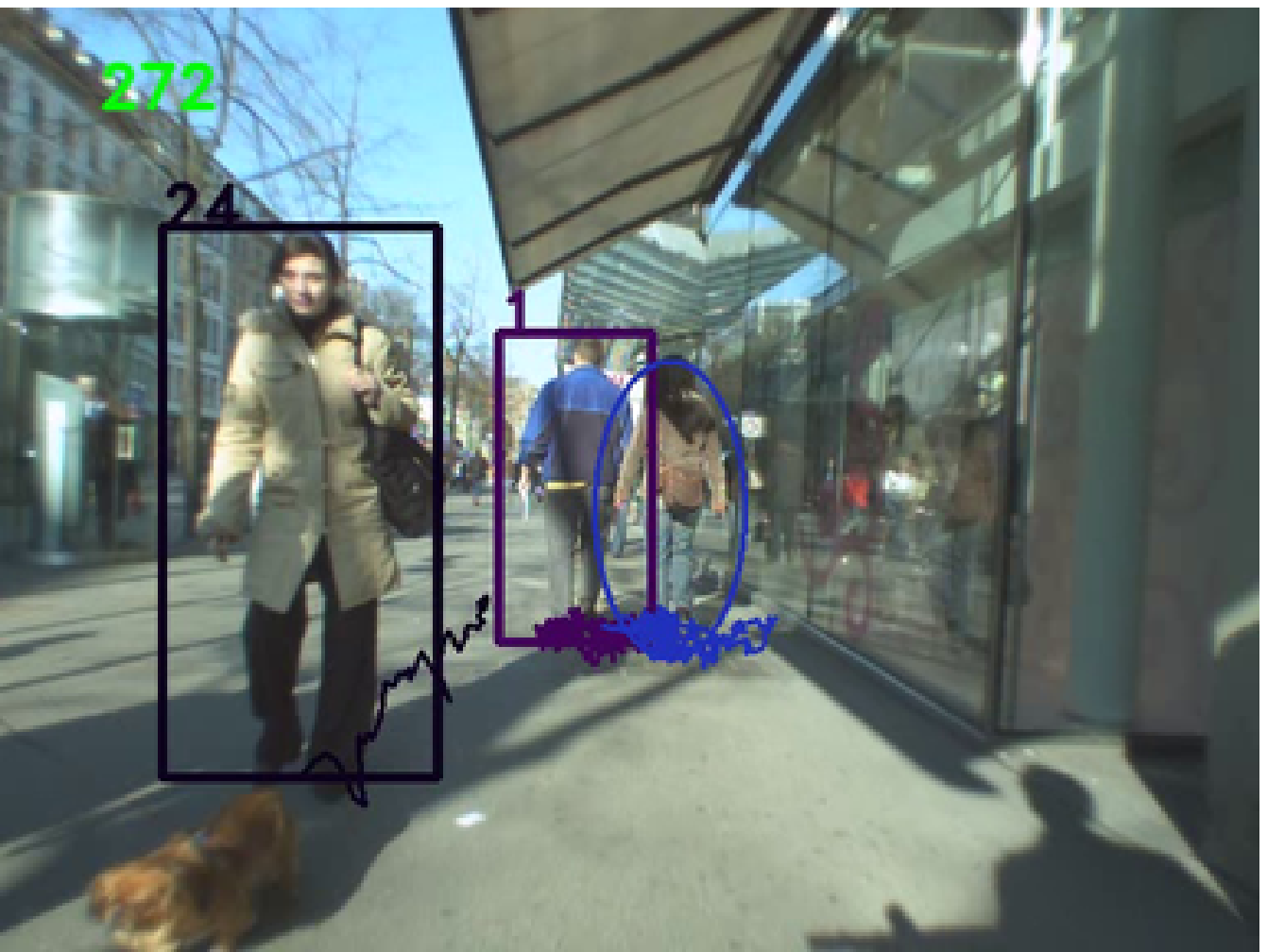} \\
    (3a) & (3b) & (3c) & (3d) & (3e) \\
    \includegraphics[scale=0.22]{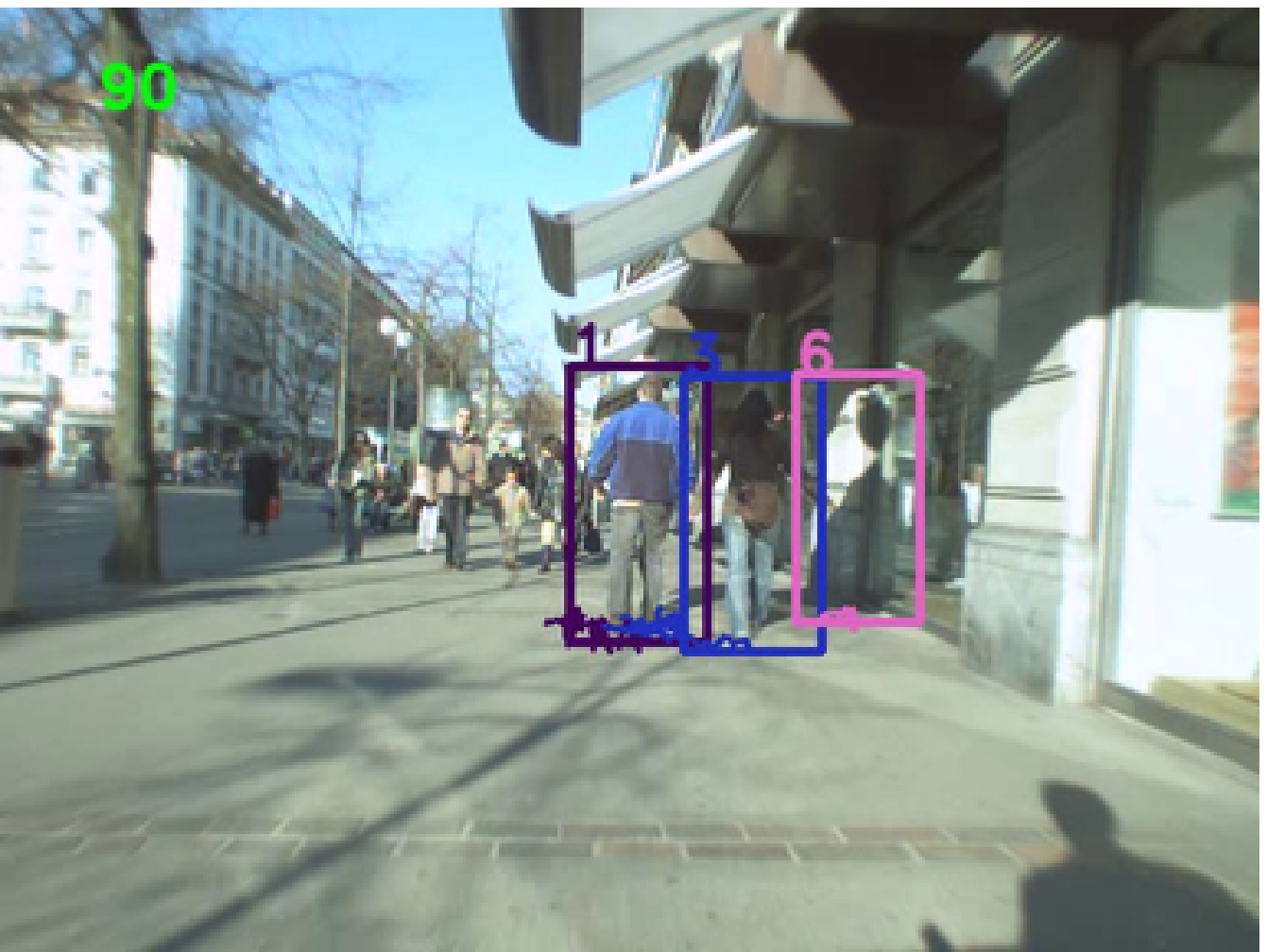} & 
    \includegraphics[scale=0.22]{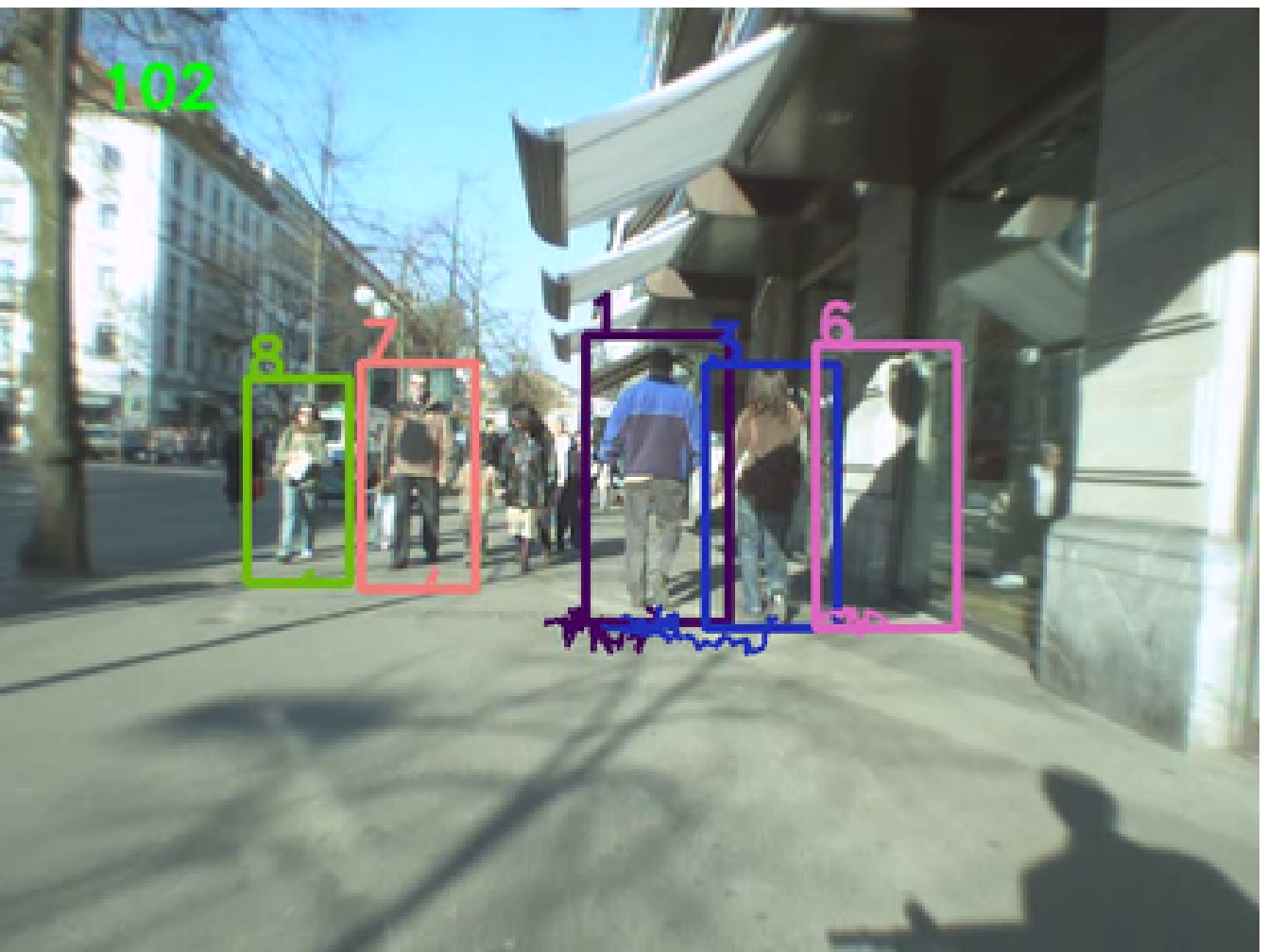} & 
    \includegraphics[scale=0.22]{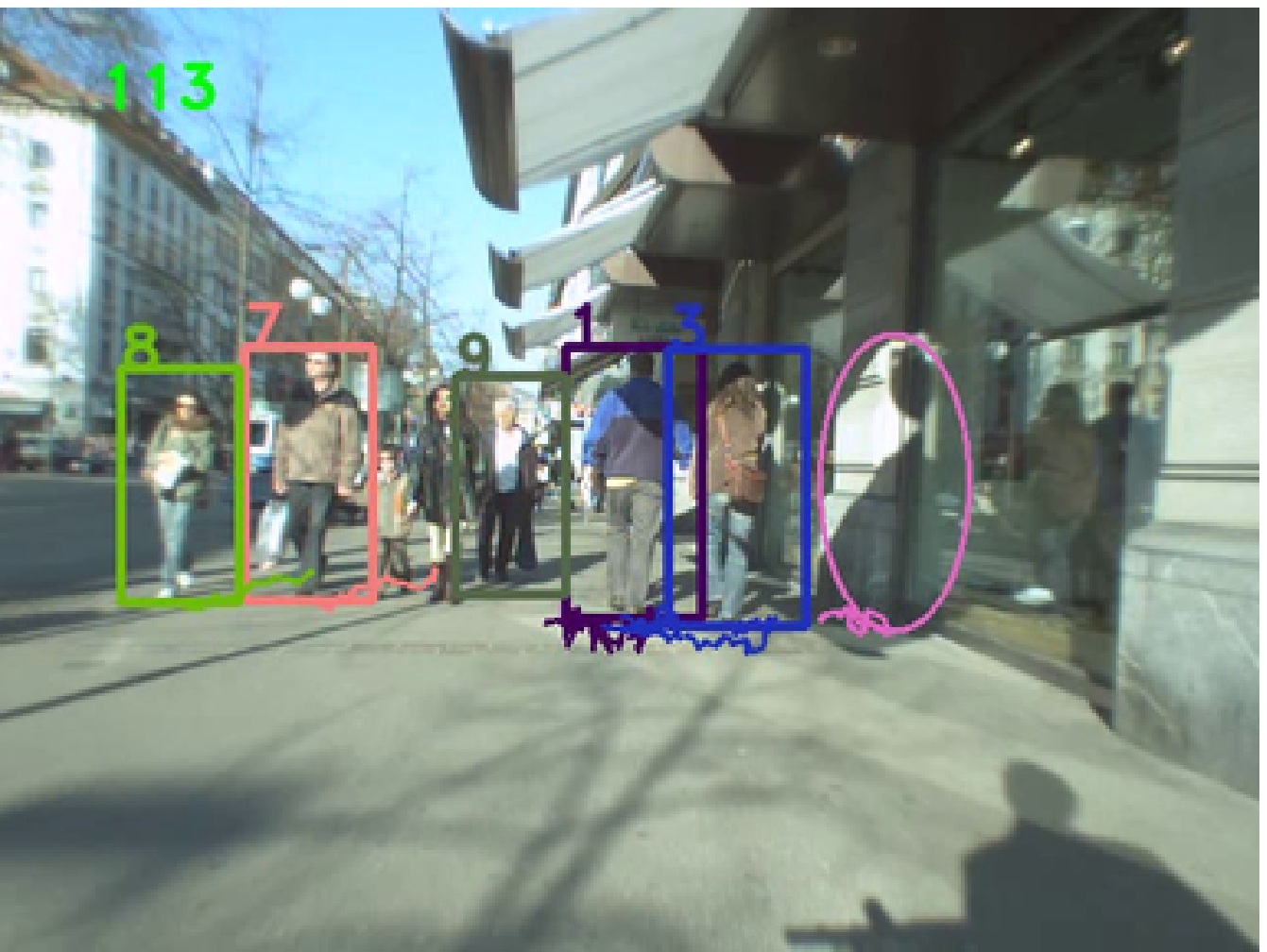} & 
    \includegraphics[scale=0.22]{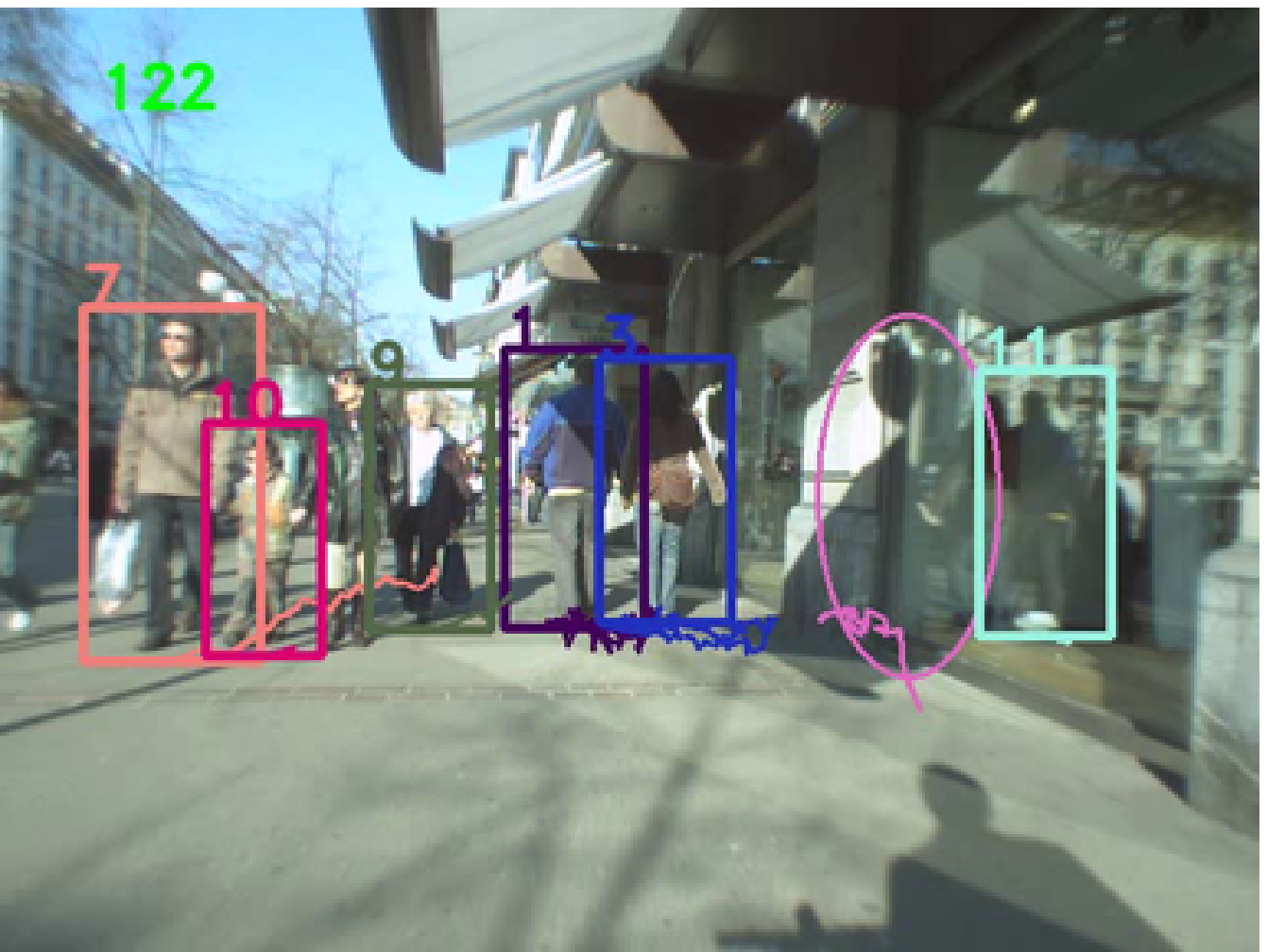} & 
    \includegraphics[scale=0.22]{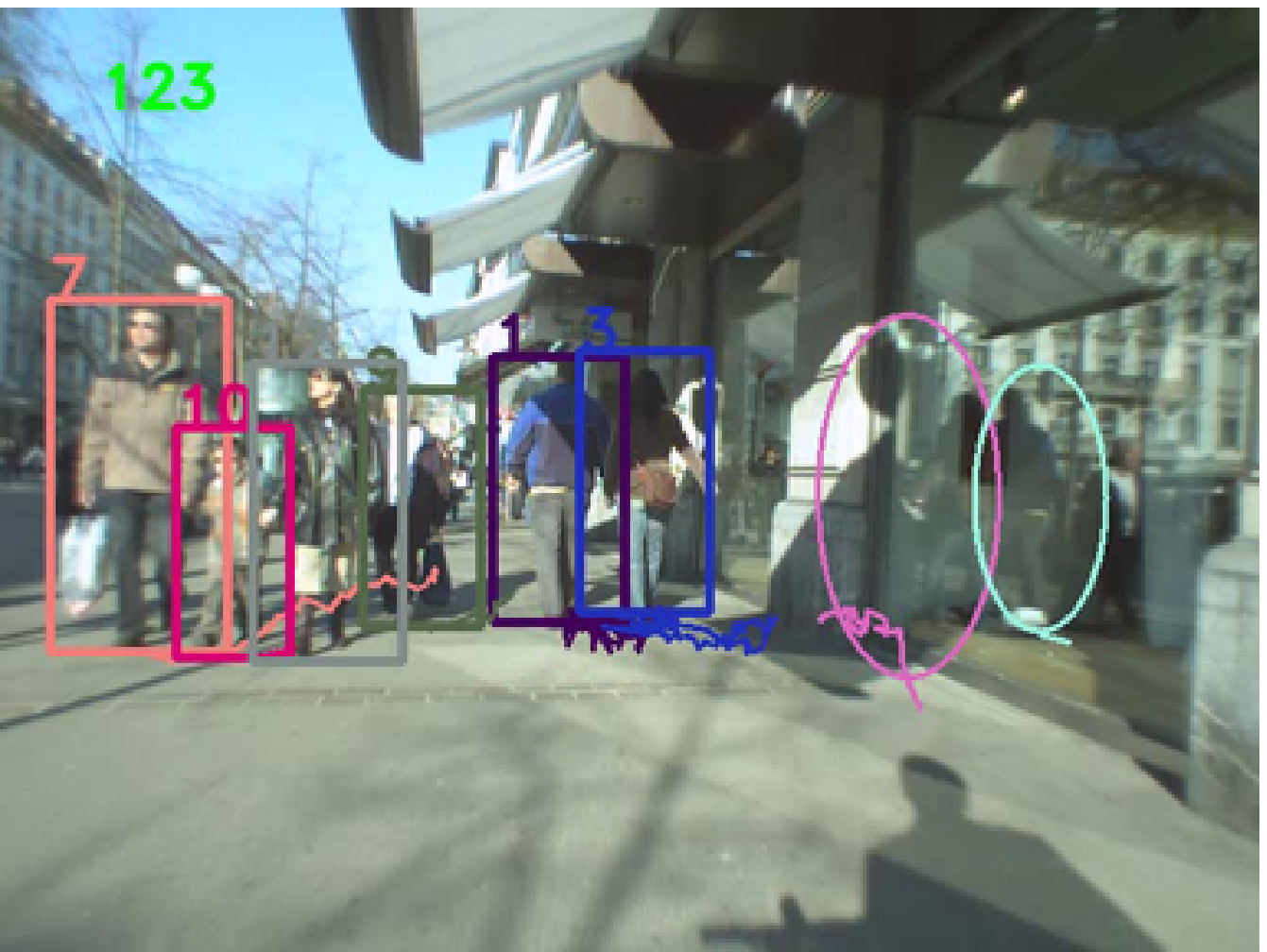} \\
    (3f) & (3g) & (3h) & (3g) & (3i) \\
    \multicolumn{5}{c}{\parbox{0.9\textwidth}{ETH3: (3a-3e) Agents 1
    and 3 are tracked successfully in presence of various false
  positives due to shadow and reflection. (3f-3j) Shows two false
trajectories generated as a result of detection failure (agent 6 and
11). }}\\
  \end{tabular}
  \caption{Snapshots of trajectories generated for various agents for
  three different ETH datasets. Predicted agent location is shown as
an ellipse. Each detected agent window is shown with a rectangular
bounding box with its agent ID. It shows several instances where the
ID switch is prevented and the target is tracked successfully despite
occlusion and other effects. }
  \label{fig:snaps} 
\end{figure*}

\begin{figure*}[!t]
  \centering
  \ContinuedFloat
  \begin{tabular}{ccccc}
    \includegraphics[scale=0.20]{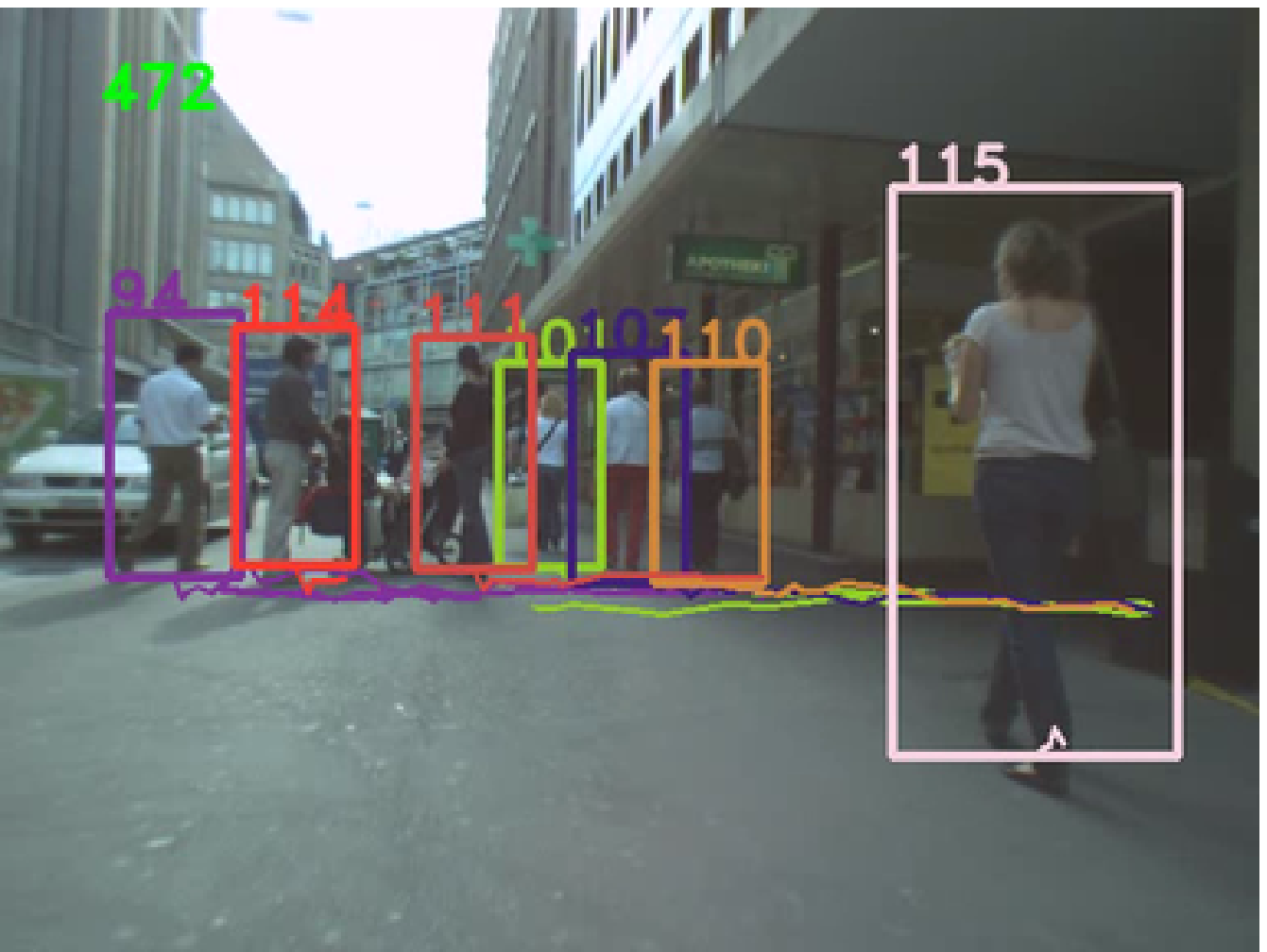} & 
    \includegraphics[scale=0.20]{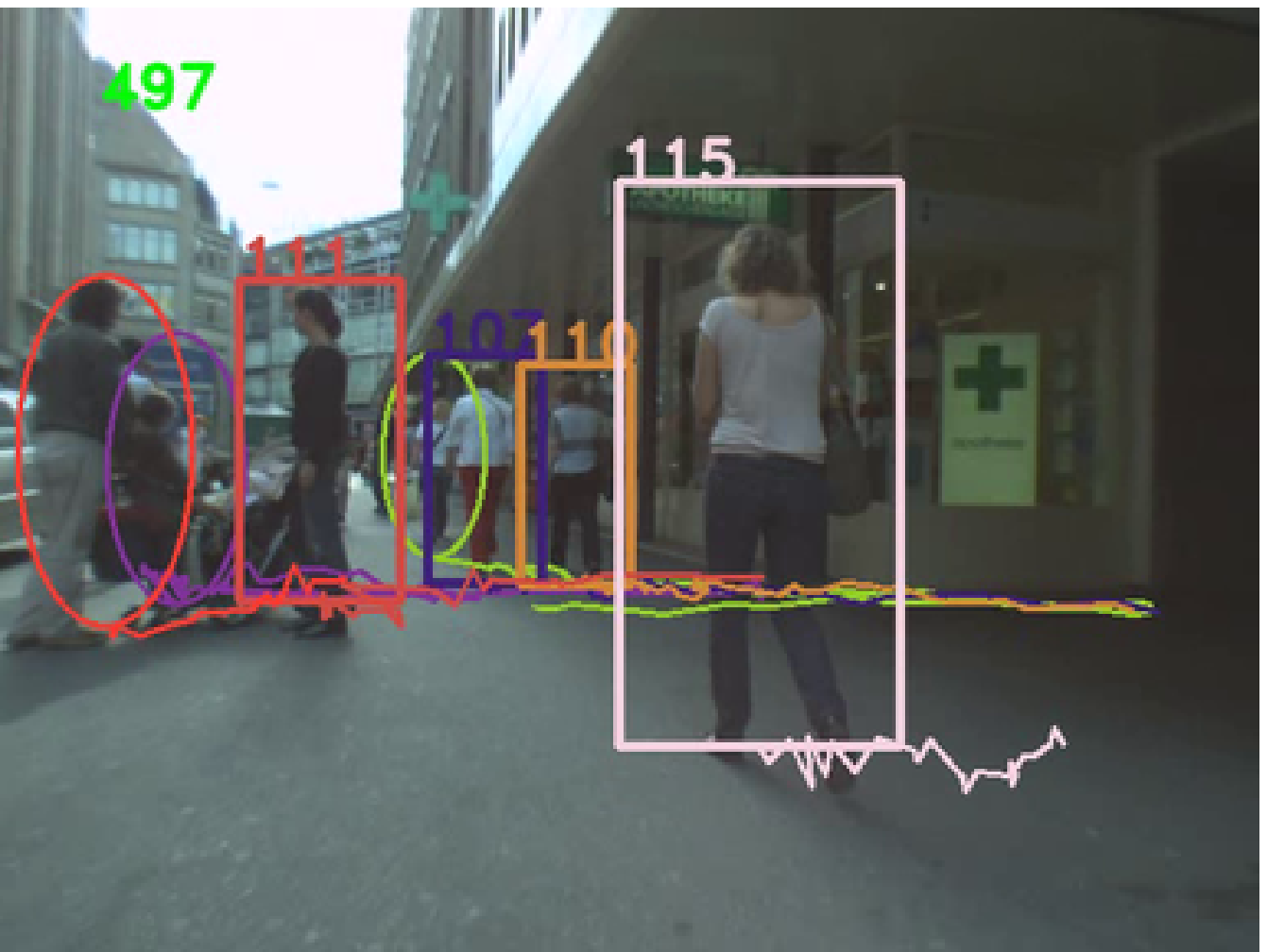} & 
    \includegraphics[scale=0.20]{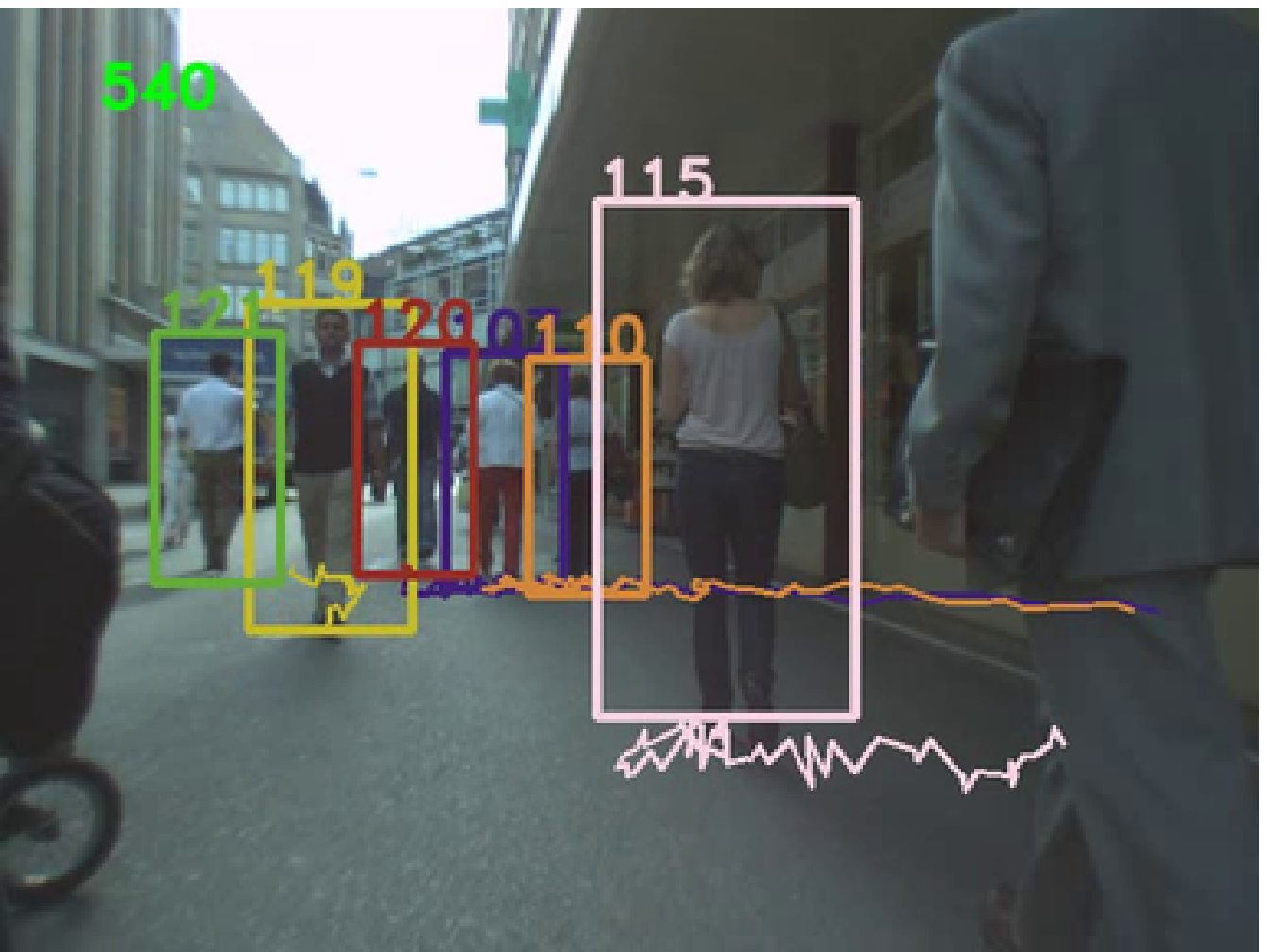} & 
    \includegraphics[scale=0.20]{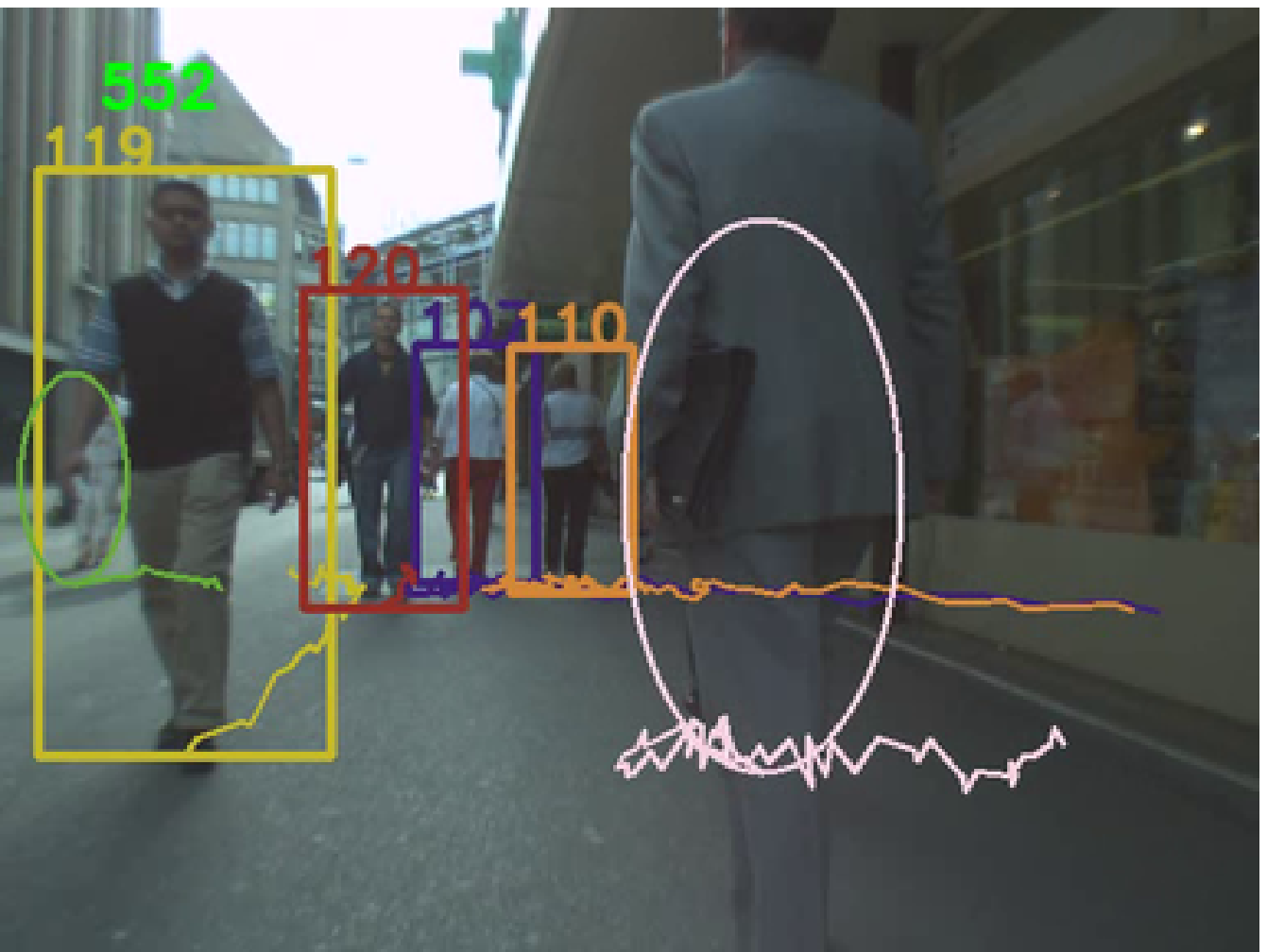} & 
    \includegraphics[scale=0.20]{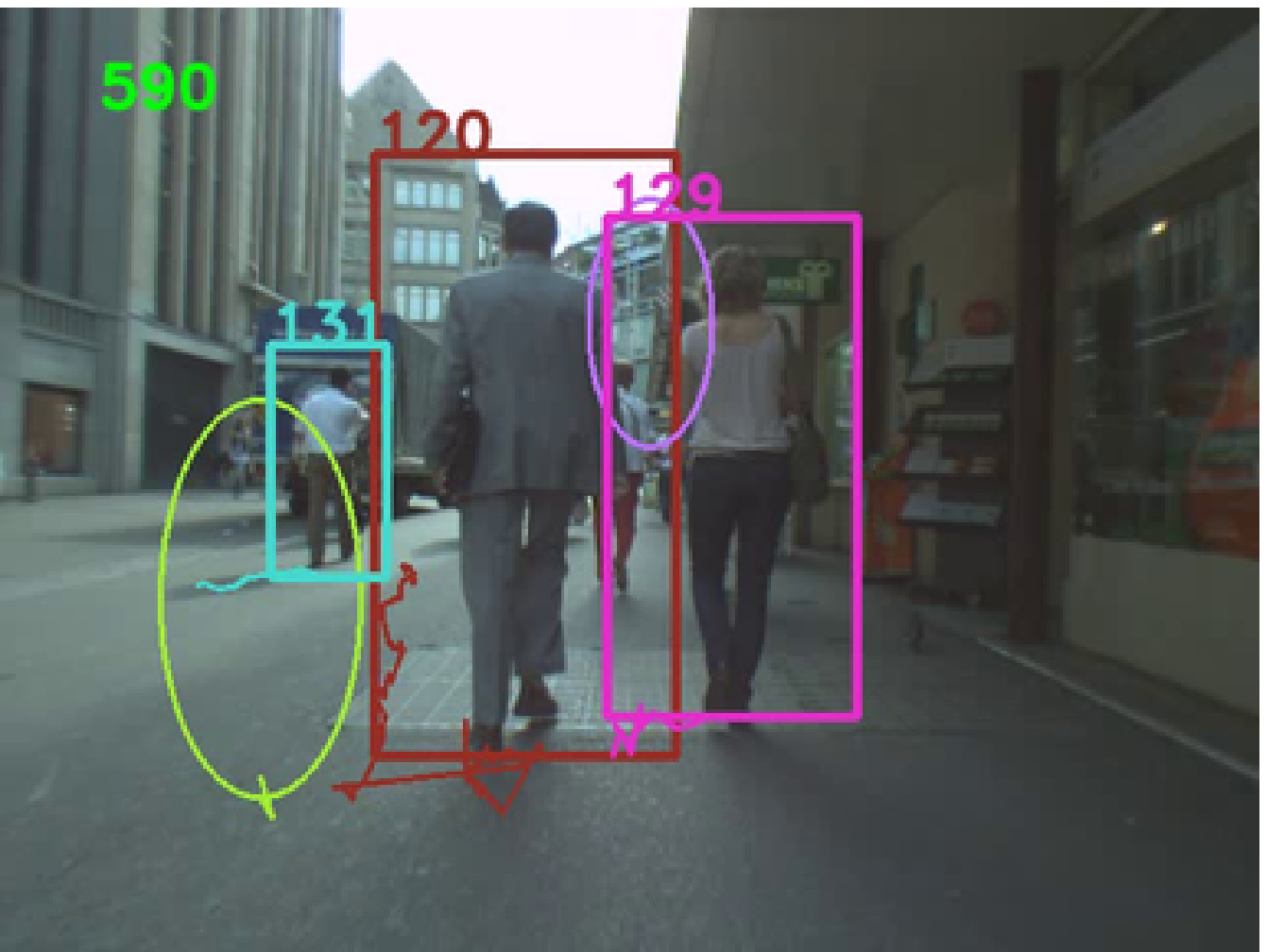} \\
    (4a) & (4b) & (4c) & (4d) & (4e) \\
    \includegraphics[scale=0.20]{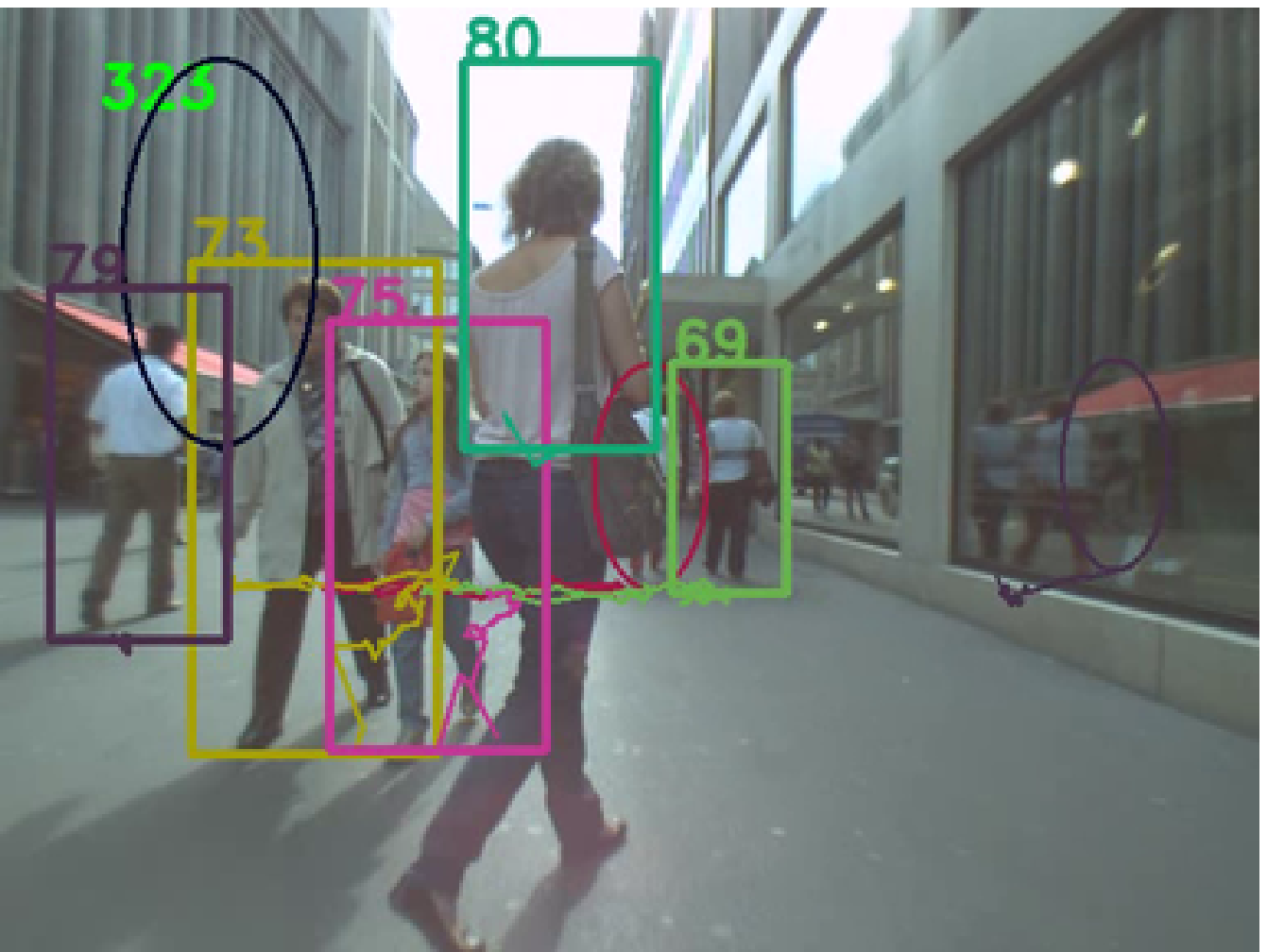} & 
    \includegraphics[scale=0.20]{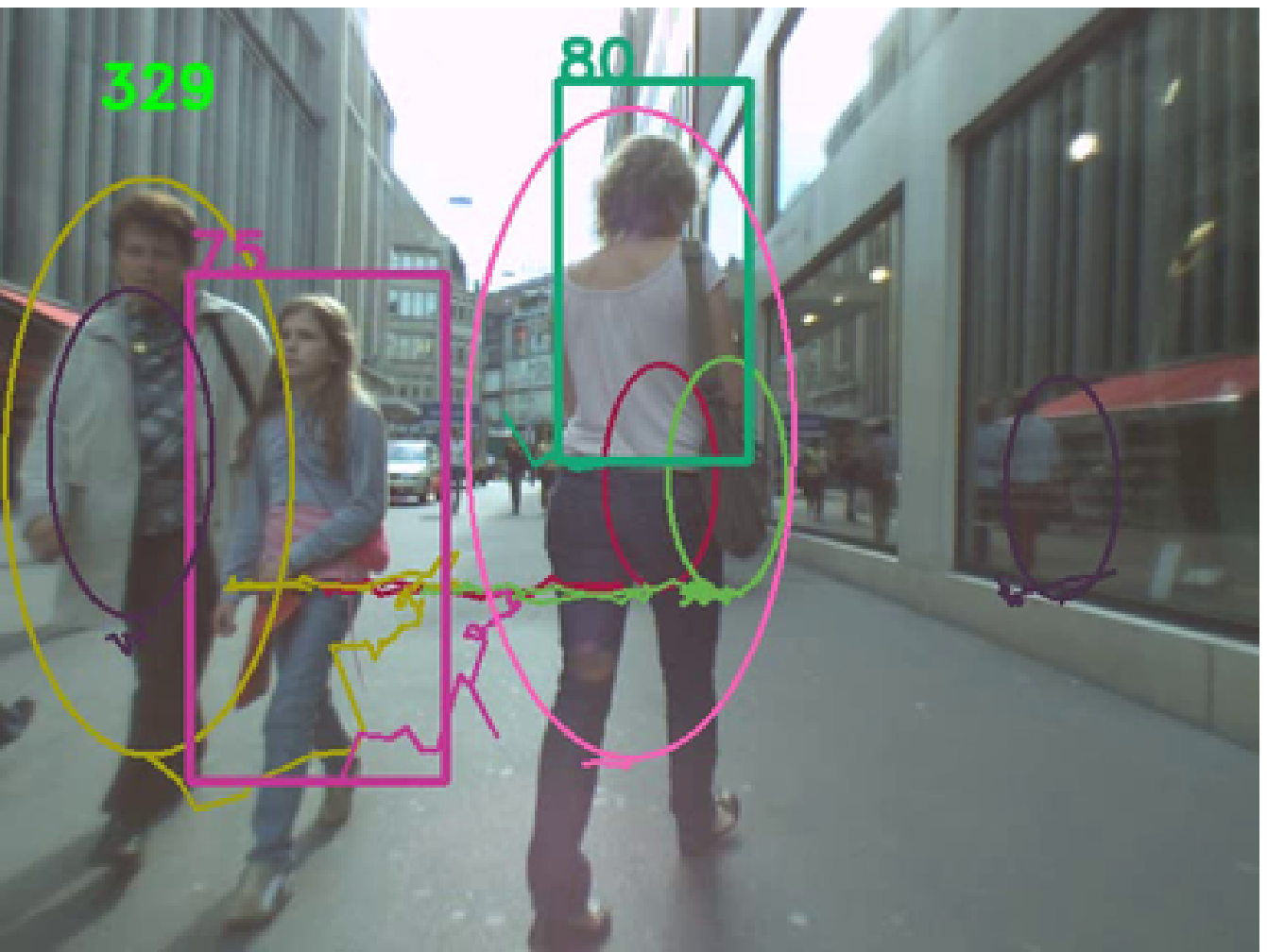} & 
    \includegraphics[scale=0.20]{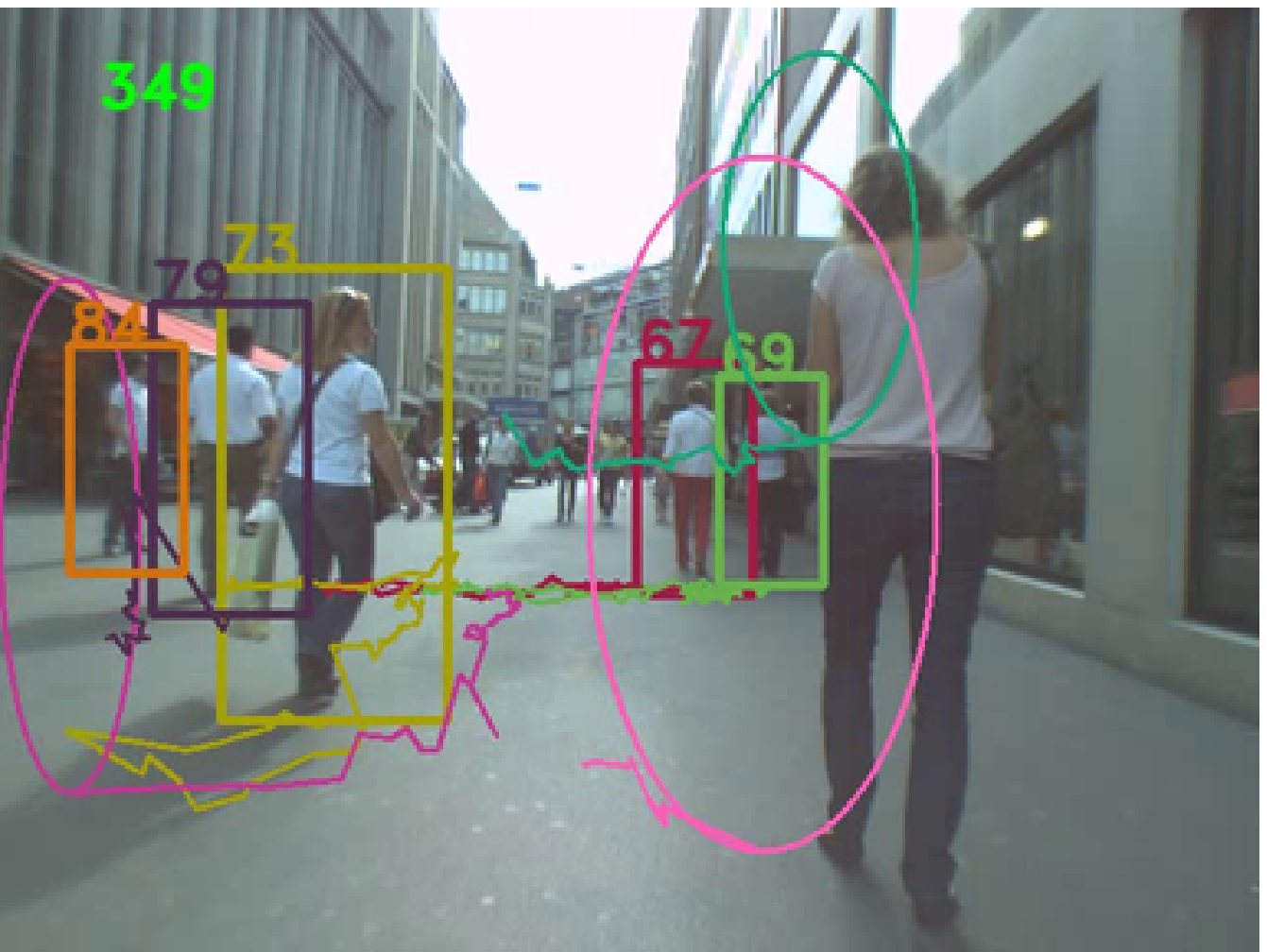} & 
    \includegraphics[scale=0.20]{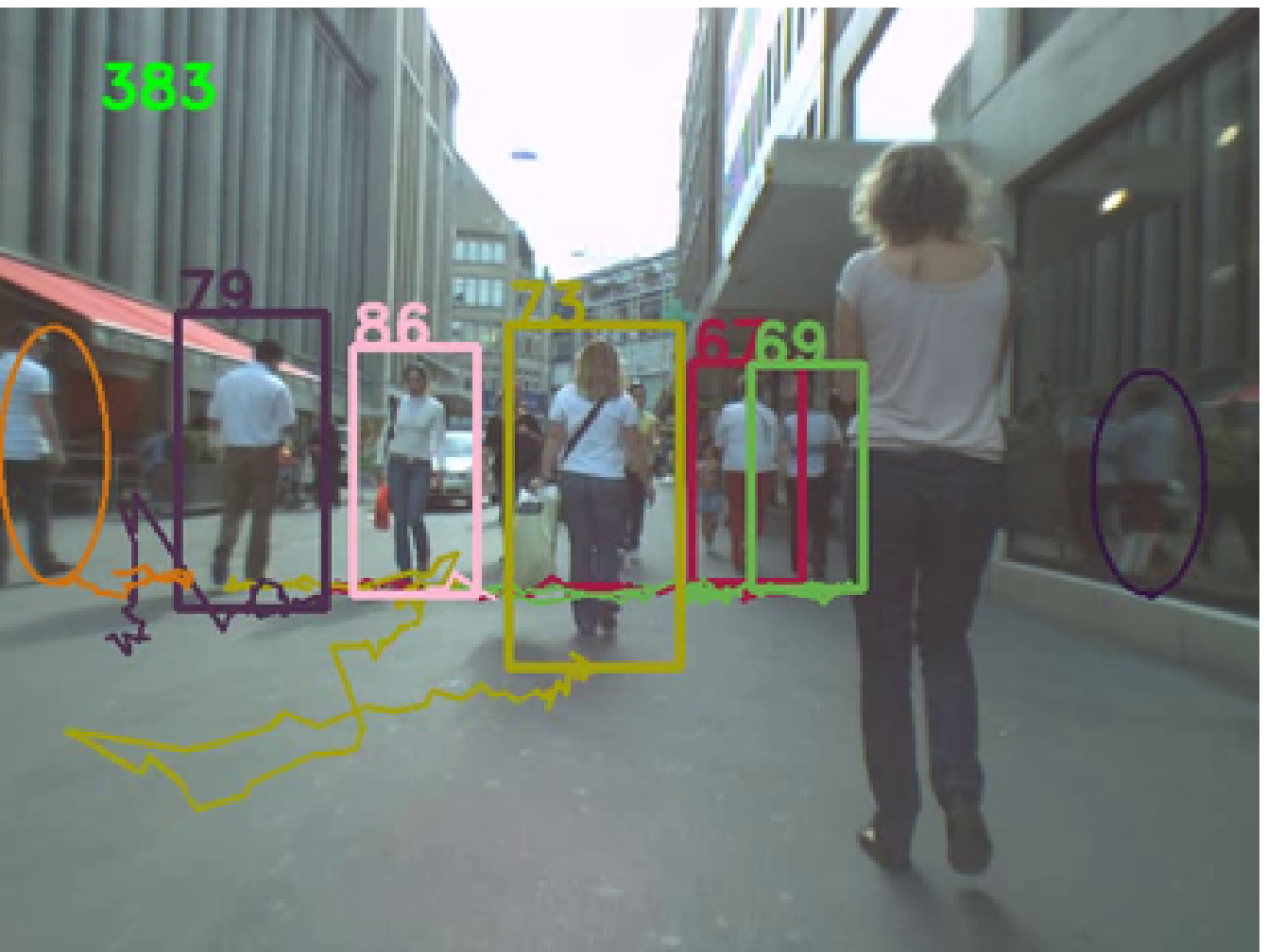} & 
    \includegraphics[scale=0.20]{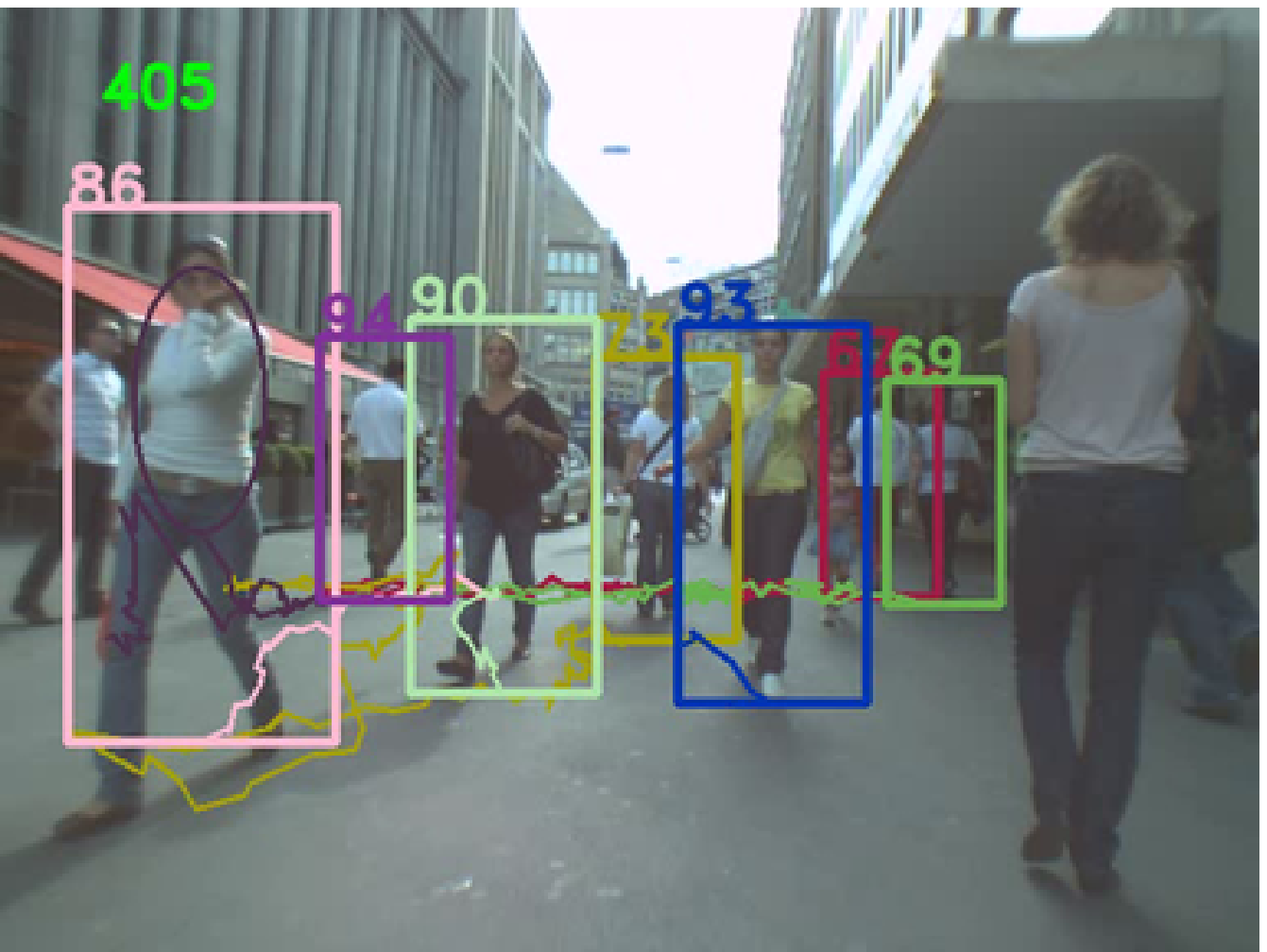} \\
    (4f) & (4g) & (4h) & (4i) & (4j) \\
    \multicolumn{5}{c}{\parbox{0.8\textwidth}{ETH4: (4a-4e) The lady
      on right gets a new ID as it recovers from occlusion. It also
      shows several cases of ID switch.  (4f-4j) Agent 79 (navy blue)
      is tracked successfully over a span of more than 100 frames and
      then undergoes an ID switch. This video has significant camera
    motion which resulting in poor tracking performance}.  }\\
    \includegraphics[scale=0.30]{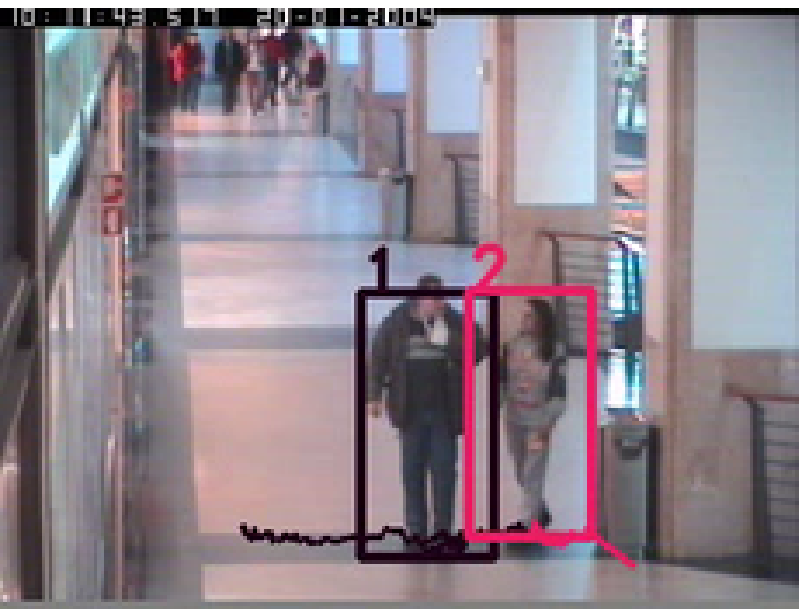} & 
    \includegraphics[scale=0.30]{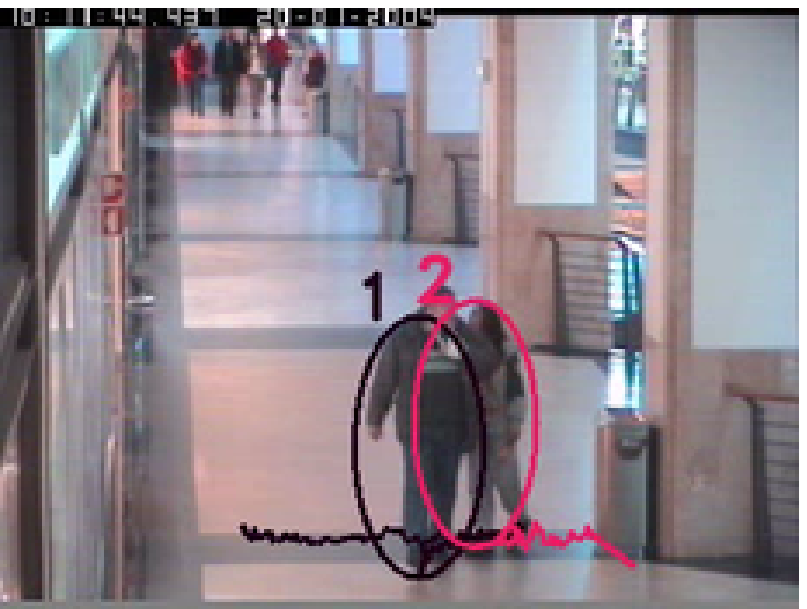} & 
    \includegraphics[scale=0.30]{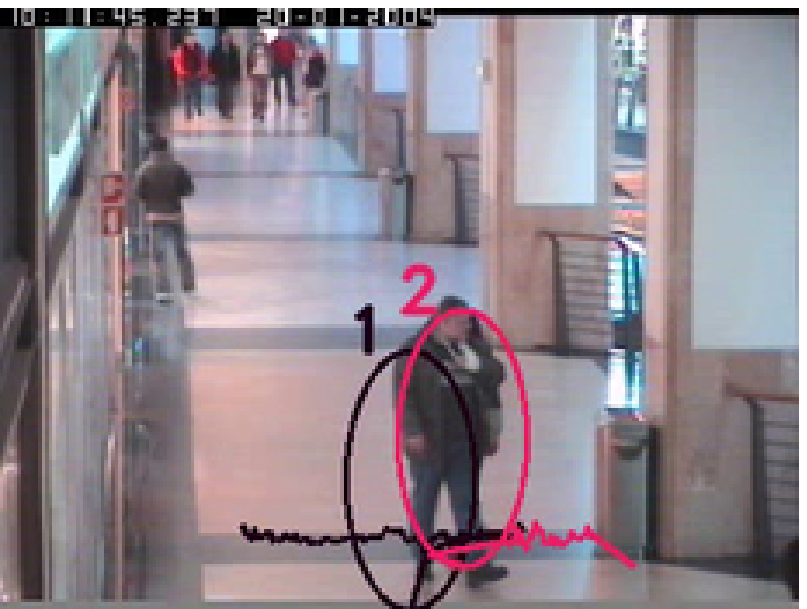} & 
    \includegraphics[scale=0.30]{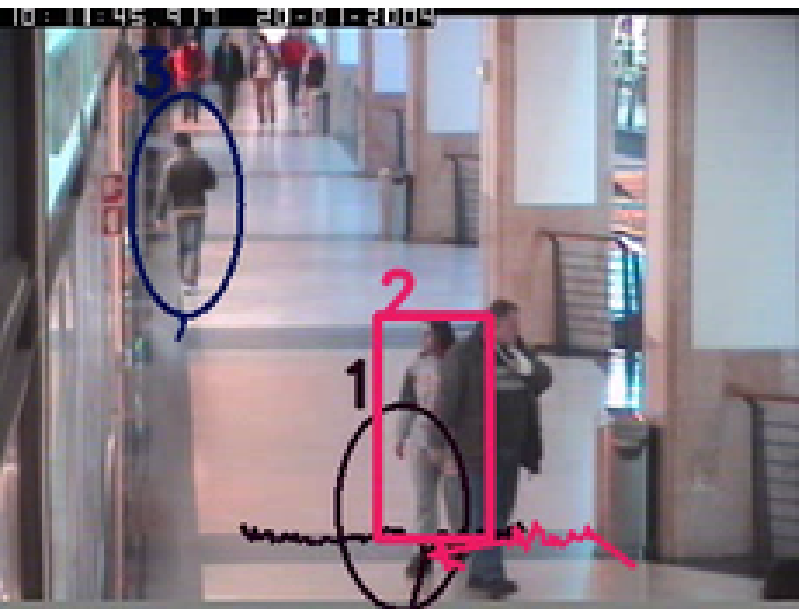} & 
    \includegraphics[scale=0.30]{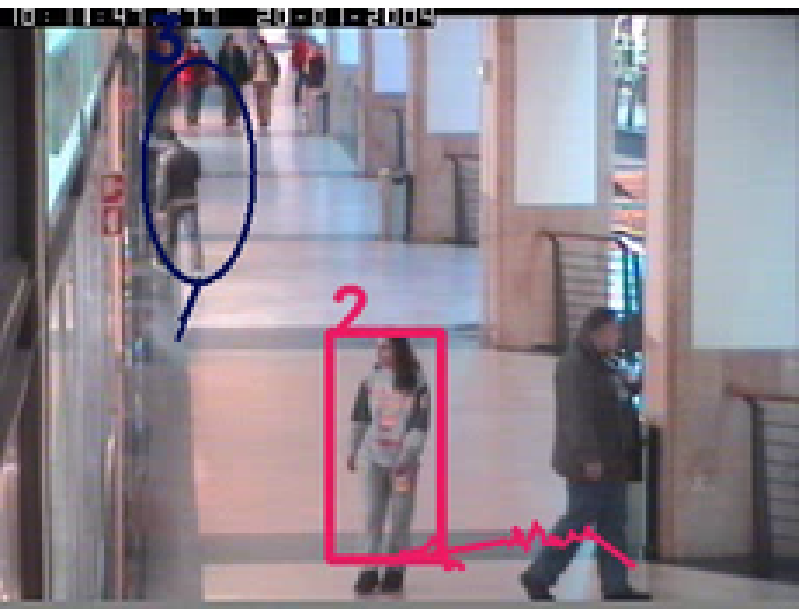} \\
    (5a) & (5b) & (5c) & (5d) & (5e) \\
    \includegraphics[scale=0.30]{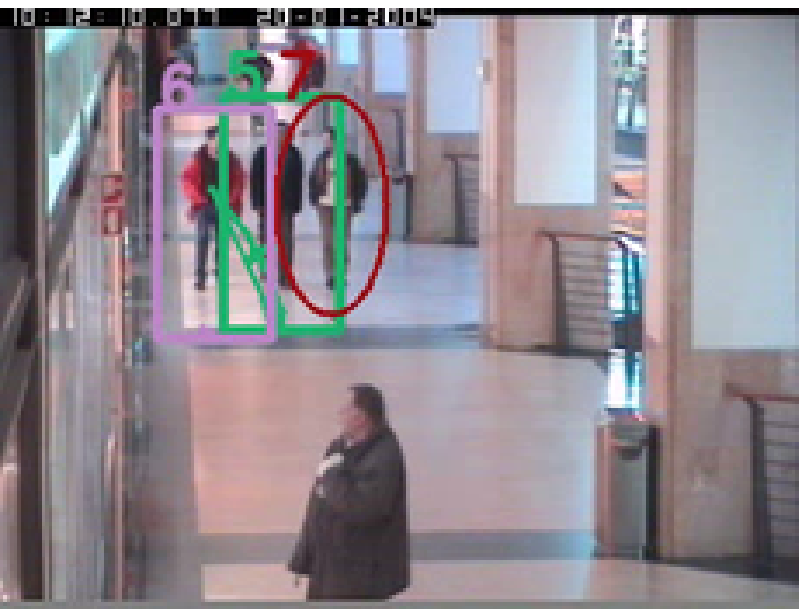} & 
    \includegraphics[scale=0.30]{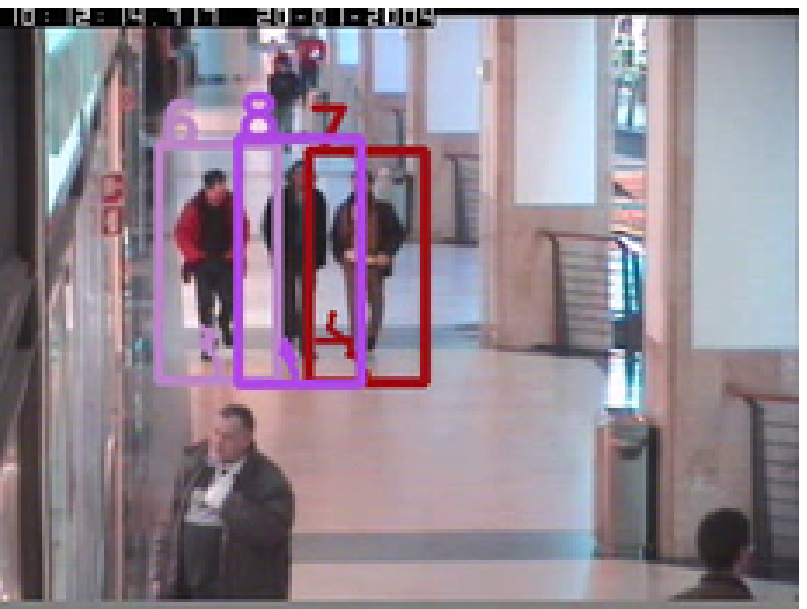} & 
    \includegraphics[scale=0.30]{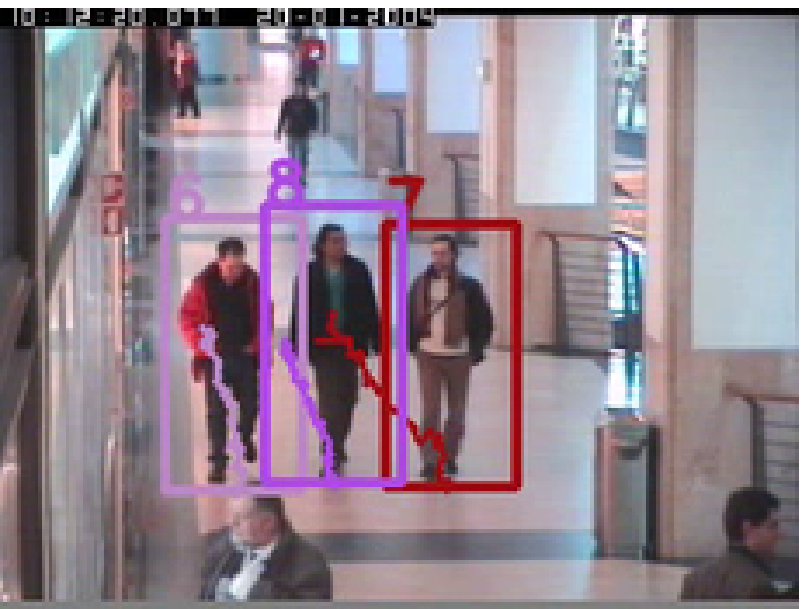} & 
    \includegraphics[scale=0.30]{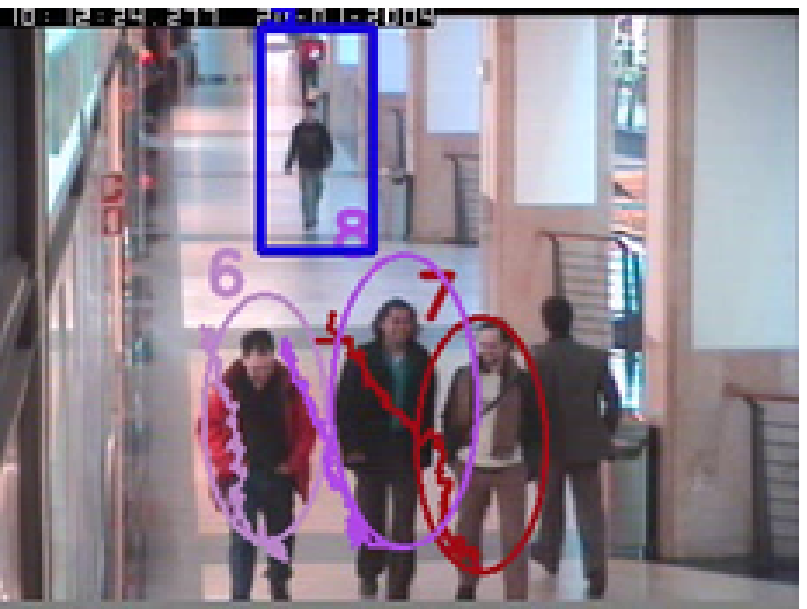} & 
    \includegraphics[scale=0.30]{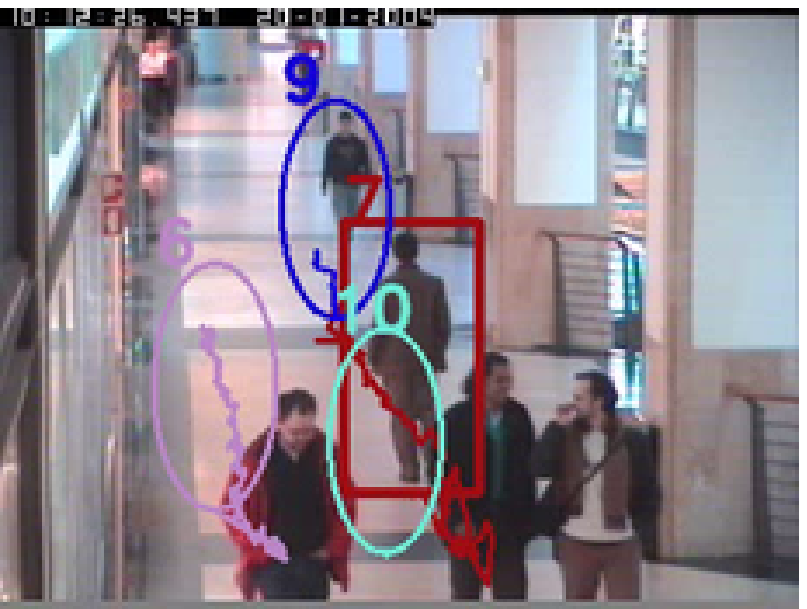} \\
  \multicolumn{5}{c}{\parbox{0.8\textwidth}{CAVIAR1: (5a-5e) Identity
  of agent 2 (shown in pink) is restored once it recovers from
occlusion by agent 1. (5f-5j) Two ID switch occurs among the three
agents which move in a group. }} 
  \end{tabular}
  \caption{Snapshots of tracking performance for Video datasets ETH4
  and CAVIAR1. CAVIAR dataset has a static background while ETH
datasets have dynamic background.}
  \label{fig:snaps2}
\end{figure*}

\section{Conclusion and Future Work} \label{sec:conc}

In this paper, we take a relook at the multi-target tracking problem.
Our main contribution lies in proposing an occlusion reasoning scheme
to solve the association among the detected agents on a frame-by-frame
basis. The scheme defines an \emph{affinity matrix} that depicts the
closeness between the estimated agent windows of the previous frame
with those obtained from a detector in the current frame. This
affinity matrix is later used by a binary integer programming (BIP)
module to find unique associations between these pair of windows. A
second stage of verification based on SURF-matching is employed to
deal with the wrong associations generated by the BIP module. This
module makes use of past agent pair information to resolve the agent
identities in the current frame. The performance of our algorithm is
compared with the latest work in this field. It is still an initial
work with a lot of scope for improvement. The work presented here will
be useful for students and practicing engineers who would like to
understand the process and the underlying challenges of the problem.

\balance
\bibliographystyle{IEEEtran}
\bibliography{ref}  % sigproc.bib is the name of the Bibliography in this case

\end{document}